\PassOptionsToPackage{numbers,sort&compress}{natbib}
\documentclass{article}

\usepackage[preprint]{neurips_2026}

\usepackage[utf8]{inputenc}
\usepackage[T1]{fontenc}
\usepackage{hyperref}
\usepackage{url}
\usepackage{booktabs}
\usepackage{amsfonts}
\usepackage{amsmath}
\usepackage{amssymb}
\usepackage{nicefrac}
\usepackage{microtype}
\usepackage{xcolor}
\usepackage{graphicx}
\usepackage{float}
\usepackage{multirow}

\newcommand{\kinetixfig}[1]{#1}
\newcommand{\liberofig}[1]{#1}
\newcommand{\liberoablationfig}[1]{#1}

\title{Understanding Asynchronous Inference Methods for Vision-Language-Action Models}

\author{%
  Ayoub Agouzoul\\
  \'Ecole polytechnique \\
  Systems Group, ETH Z\"urich \\
  \texttt{ayoub.agouzoul@polytechnique.edu} \\
}

\begin{document}

\maketitle

\begin{abstract}
Vision-Language-Action (VLA) models offer a promising path to generalist robot control, but their inference latency causes observation staleness when generated actions are executed asynchronously. Several methods have been proposed concurrently to mitigate this problem: inference-time inpainting (IT-RTC), training-time delay simulation (TT-RTC), future-state-aware conditioning (VLASH), and lightweight residual correction (A2C2). Each takes a fundamentally different approach, but they have so far been evaluated independently with different codebases, base policies, and protocols. We present a systematic comparison of these four methods under controlled conditions. We develop two unified codebases that integrate all methods with harmonized library and dataset versions, and we benchmark them on the Kinetix suite with MLPMixer policies and on the LIBERO manipulation benchmark with SmolVLA, sweeping inference delays up to $d=20$ control steps. A2C2's per-step residual correction is the most effective method on Kinetix, holding above 90\% solve rate up to $d{=}8$, and also leads on LIBERO from $d{\geq}4$ onwards. IT-RTC is competitive at low delays but degrades sharply under long chunks ($H{=}30$) and high delays. TT-RTC is the most robust training-based method: stable across $d_{\max}$ choices, generalizes beyond its training delay distribution, and adds zero inference overhead. VLASH exhibits a clear low-delay vs.\ high-delay trade-off governed by the fine-tuning delay range $[0,d_{\max}]$. Code is available at \url{https://github.com/TheAyos/async-vla-inference}.
\end{abstract}

\section{Introduction}\label{sec:introduction}

Vision-Language-Action (VLA) models have emerged as a powerful paradigm for generalist robot control, mapping visual observations and natural-language instructions to robot actions~\citep{rt1,openvla,pi0,pi05,smolvla}. A key design choice in modern diffusion- and flow-matching-based VLA policies is \emph{action chunking}~\citep{aloha}: instead of predicting one action per observation, the policy predicts a sequence of $H$ future actions in a single forward pass, reducing compounding errors and amortizing inference cost. However, action chunking interacts poorly with the high inference latency of large VLA models---typically tens to hundreds of milliseconds~\citep{runningvlas,vlaperf}. By the time a chunk is ready, the robot's state and environment have already moved on, so executed actions are conditioned on a stale observation and may no longer be optimal.

\emph{Synchronous} execution avoids this problem by halting the robot during inference, but caps the control frequency at the policy's end-to-end inference rate, which is impractical for tasks requiring fast reaction times. \emph{Naive asynchronous} execution maintains a high control rate by decoupling inference from execution and switching to the new chunk as soon as it is available~\citep{smolvla}, but introduces jitter and discontinuities at chunk boundaries that visibly degrade task performance~\citep{itrtc}. Four methods have recently been proposed to bridge this gap, each targeting a different failure mode:
\begin{itemize}
    \item \textbf{IT-RTC}~\citep{itrtc}: training-free flow-matching guidance that inpaints fresh actions onto a partially frozen chunk at inference time.
    \item \textbf{TT-RTC}~\citep{ttrtc}: training-time delay simulation that conditions the policy on action prefixes so it learns to predict only the non-stale postfix, with no inference overhead.
    \item \textbf{VLASH}~\citep{vlash}: estimates the robot's future state at execution time using known previous actions, and conditions the policy on this rolled-forward state.
    \item \textbf{A2C2}~\citep{a2c2}: a lightweight correction head that runs at every control step, producing residual adjustments to the base policy's stale actions.
\end{itemize}

Despite their shared motivation, these methods have been developed and evaluated in incompatible codebases, on different benchmarks, and with different base policies. This makes direct comparison difficult and obscures the trade-offs between them.

\paragraph{Contributions.} (i)~Two \textbf{unified codebases} (\texttt{bt-kinetix} in JAX/Flax NNX, \texttt{bt-libero} in PyTorch/LeRobot) that integrate all four methods with harmonized libraries, datasets, and base policies. (ii)~A \textbf{systematic comparison} on the Kinetix~\citep{kinetix} continuous-control suite (10 dynamic environments, $H \in \{16, 30\}$, delays up to $d{=}15$) and on the LIBERO~\citep{libero} manipulation benchmark with SmolVLA (delays up to $d{=}20$). (iii)~\textbf{Practical insights} on training cost, inference overhead, and out-of-distribution delay generalization that inform the choice of asynchronous inference strategy for future VLA design and deployment.

\section{Background and Methods}\label{sec:background}

\subsection{The Asynchronous Inference Gap}\label{sec:gap}

Following~\citet{itrtc}, we define the inference delay $d = \lfloor \delta / \Delta t \rfloor$, where $\delta$ is the wall-clock inference time and $\Delta t$ is the control period. At time $t$ the robot captures observation $o_t$ and submits it to the policy. During the $d$ control steps that inference takes, the robot must continue to act and so executes the tail of the previous chunk. When the new chunk $A_t = \pi(o_t)$ is ready at step $t{+}d$, its first $d$ entries correspond to elapsed timesteps and are obsolete; the robot begins executing from $a_{t+d}$ onward. Naive asynchronous execution thus suffers from three interrelated failure modes: \emph{chunk-boundary discontinuity} (the new chunk is incompatible with the just-committed actions), \emph{observation staleness} (the new chunk is conditioned on a $d$-step-old observation), and \emph{open-loop degradation} (committed actions cannot adapt during the chunk's lifetime).

\subsection{Four Methods, Four Failure Modes}\label{sec:methods}

The four methods we compare each primarily target one of these failure modes; their workings are detailed in Appendix~\ref{app:method_details}.

\paragraph{IT-RTC~\citep{itrtc}} addresses chunk-boundary discontinuity by treating the first $d$ actions of a new chunk as known constraints and \emph{inpainting} the remaining $H{-}d$ actions during flow-matching denoising, drawing on diffusion-based image inpainting~\citep{pokle2024trainingfreelinearimageinverses}. Each Euler step adds a pseudoinverse-guidance term~\citep{song2023pseudoinverseguided} that steers the denoised actions toward satisfying the prefix constraint, with a soft exponentially decaying mask blending frozen and free actions. IT-RTC requires no retraining but invokes a backward (VJP) pass per denoising step, tripling the denoising-loop FLOPs (Section~\ref{sec:cost}).

\paragraph{TT-RTC~\citep{ttrtc}} attacks the same problem at training time. Through three minimal architecture changes---per-token flow timesteps via adaLN-Zero~\citep{dit}, a non-noisy ground-truth prefix during training, and a masked loss that fires only on the postfix---the policy learns to predict postfix actions conditioned on a committed prefix. The training-time delay $d$ is sampled with exponential decay $p(d) \propto \lambda^d$ for $d \in \{0, \ldots, d_{\max}\}$. At inference, TT-RTC is a drop-in replacement for IT-RTC with zero overhead.

\paragraph{VLASH~\citep{vlash}} directly targets observation staleness by modifying the \emph{state conditioning}. When inference begins on $o_t$, VLASH rolls the proprioceptive state forward by $\Delta$ steps using the known committed actions, and conditions the (otherwise unmodified) policy on $(o_t, \hat{s}_{t+\Delta})$ rather than $(o_t, s_t)$. A temporal-offset training augmentation samples $\delta \in \{0, \ldots, \Delta_{\max}\}$ and trains the policy to predict the action chunk starting at $t{+}\delta$ from the original observation $o_t$ but the shifted state $s_{t+\delta}$, encouraging reliance on the (varying) state rather than the (fixed) image.

\paragraph{A2C2~\citep{a2c2}} targets open-loop degradation. Rather than modify the base policy, it trains a lightweight correction head $\pi_{\text{A2C2}}$ that runs at every control step and emits a residual $\Delta a_{t+k}$ added to the base action $a_{t+k}^{\text{base}}$. The head conditions on the latest observation, base-policy hidden features, language, and a sinusoidal time-in-chunk feature $\tau_k = (\sin(2\pi k / H), \cos(2\pi k / H))$ that signals how stale the current base action is. The head is small enough to run inline within the control loop and the base policy is left frozen, so A2C2 is in principle composable with the other three methods.

\subsection{Mapping Inference Delay to Real-World Deployments}\label{sec:vlaperf}

Prior work on asynchronous chunking~\citep{itrtc,ttrtc,vlash} has primarily evaluated delays up to $d{=}4$. At a 50\,Hz control rate ($\Delta t = 20$\,ms, the standard for the $\pi_0$ family~\citep{pi0,pi05}), this corresponds to only 80\,ms and does not capture the regime relevant to many real deployments. Drawing on the roofline-based latency analysis of \citet{vlaperf} for $\pi_0$, Table~\ref{tab:delay_mapping} maps representative deployment scenarios to their expected delay. Even an on-device Jetson Thor incurs $d{=}2$, an edge server over 4G already reaches $d{=}3$, and cloud inference over constrained networks pushes $d$ to $13$ or beyond. These numbers motivate evaluating delays well past $d{=}4$ and choosing $d_{\max}$ for training-based methods accordingly.

\begin{table}[h]
\centering
\caption{Inference delay $d$ at 50\,Hz control ($\Delta t = 20$\,ms) for representative $\pi_0$ deployment scenarios, using the latency estimates of~\citet{vlaperf}. $d = \lfloor \delta / \Delta t \rfloor$.}
\label{tab:delay_mapping}
\small
\begin{tabular}{@{} l l r r @{}}
\toprule
\textbf{Scenario} & \textbf{Configuration} & \textbf{Latency $\delta$} & \textbf{Delay $d$} \\
\midrule
Datacenter GPU (local)  & B100, no network             & 3.2\,ms   & 0  \\
Consumer GPU (local)    & RTX 4090, no network         & 31.1\,ms  & 1  \\
Edge GPU (on-device)    & Jetson Thor, no network      & 52.6\,ms  & 2  \\
Edge server (Wi-Fi 7)   & B100 + Wi-Fi 7               & 8.4\,ms   & 0  \\
Edge server (5G)        & B100 + 5G                    & 27.8\,ms  & 1  \\
Edge server (4G)        & B100 + 4G                    & 73.0\,ms  & 3  \\
Cloud server (fast)     & B100 + wired + fast cloud    & 23.4\,ms  & 1  \\
Cloud server (slow)     & B100 + 4G + slow cloud       & 273.4\,ms & 13 \\
\bottomrule
\end{tabular}
\end{table}

\section{Experimental Setup}\label{sec:setup}

\subsection{Unified Codebases}\label{sec:codebases}

A practical obstacle to comparing the four methods is that each was developed in a separate codebase with different dependencies, dataset formats, and evaluation protocols. The IT-RTC and TT-RTC papers share a Kinetix-only JAX codebase; VLASH and A2C2 each ship LIBERO-specific codebases pinned to incompatible LeRobot/PyTorch versions, plus distinct JAX implementations for Kinetix. We developed two unified codebases to remove these as confounders.

\textbf{\texttt{bt-kinetix}} (JAX, Flax NNX, Optax) integrates the flow-matching policy training, TT-RTC's prefix conditioning, VLASH's temporal-offset augmentation, and A2C2's residual training under a single entry point. All methods share the same MLP-Mixer~\citep{mlpmixer} backbone, data loader, and evaluation protocol. We additionally extended the original IT-RTC framework, which only supported $H{=}8$, to arbitrary $H \in \{8, 16, 30\}$, enabling delay sweeps up to $d{=}15$.

\textbf{\texttt{bt-libero}} (PyTorch, LeRobot v0.4.2, Accelerate) supports SmolVLA~\citep{smolvla} (and in principle $\pi_0$/$\pi_{0.5}$) under a unified evaluation harness with five modes: \texttt{baseline}, \texttt{vlash}, \texttt{tt\_rtc}, \texttt{it\_rtc}, and \texttt{a2c2}. We re-implemented the A2C2 residual pipeline (originally on LeRobot v0.2.x) and the IT-RTC inference-time guidance into the VLASH-derived training stack, harmonized dataset formats across LIBERO suites, and added multi-GPU distributed training for the A2C2 head via \texttt{accelerate}.

\subsection{Benchmarks and Protocols}\label{sec:benchmarks}

\textbf{Kinetix~\citep{kinetix}} is a JAX-based 2D physics suite for continuous control. Following~\citet{itrtc}, we use 10 of the 12 hand-designed dynamic environments (excluding \texttt{mjc\_walker} and \texttt{trampoline}, which exhibited unstable behavior during our experiments). Observations are 679-dimensional symbolic vectors. Actions control joint torques/forces. All Kinetix experiments share an MLP-Mixer base ($\sim$310K params, 4 layers, $C{=}256$, 5 Euler steps), trained per the four-phase pipeline of~\citet{itrtc,ttrtc} (PPO experts $\to$ data collection $\to$ flow-matching base $\to$ method-specific fine-tune); see Appendix~\ref{app:hyperparams}. We evaluate two chunk sizes, $H{=}16$ and $H{=}30$, with 2048 parallel rollouts per configuration $(d,s,\text{method})$ and per environment. We sweep over delays $(d, s) = (d, \max(1,d))$ for $d \in \{0, \ldots, \lfloor H/2 \rfloor\}$ and over execution horizons $(d,s)=(1,h)$ for $h \in \{2, \ldots, H\}$. Reported metrics are solve rate (with 95\% Wilson CIs) and mean episode length (number of steps to solve).

\textbf{LIBERO~\citep{libero}} is a manipulation benchmark on a 7-DoF Franka arm with parallel-jaw gripper and dual workspace cameras (256$\times$256). We evaluate on the four LIBERO suites (Spatial, Object, Goal, 10-task; 40 tasks total), 10 rollouts per task per $(d,s)$, with $d \in \{0,1,2,4,8,15,20\}$ and $s = 50$. The base policy is SmolVLA fine-tuned from the public \texttt{smolvla\_libero} checkpoint with $d_{\max} \in \{0,4,8,16\}$ for VLASH variants and $d_{\max} \in \{4,8\}$ for TT-RTC; the baseline is the same checkpoint fine-tuned for the same number of steps with $d_{\max}{=}0$, to ensure that the checkpoints are trained on the same dataset for a comparable number of optimization steps. Only the action expert is trained, with the VLM layers frozen. The A2C2 correction model is a 6-layer Transformer ($d_{\text{model}}{=}512$, 8 heads, FFN 2048) with a frozen ResNet-18 vision backbone and VLM-feature projection (32.0M total / 20.8M trainable params). Inference delay is simulated by feeding the policy the observation from $d$ steps ago while executing the first $s$ actions of each predicted chunk; for A2C2, observations are always current and staleness is simulated at the action level. Hyperparameters are listed in Appendix~\ref{app:libero_hyperparams}.

\textbf{Important caveat for VLASH on LIBERO.} Because the LIBERO state dimension (9: end-effector pose + gripper width) differs from the action dimension (7: end-effector velocity + gripper command), state rollforward via committed actions (as on Kinetix) is not possible. Due to time constraints, we follow the original VLASH authors' implementation\footnote{See \url{https://github.com/mit-han-lab/vlash/issues/10\#issuecomment-3700627643}.} and use ground-truth future states from the demonstration dataset for $\hat{s}_{t+\delta}$. This is an information advantage that would not be available in real-world deployment. We re-emphasize this when interpreting LIBERO results (Section~\ref{sec:libero_results}).

\textbf{Hardware.} Kinetix training and evaluation, and LIBERO training, use 4$\times$ H100 NVL GPUs (CUDA 13.0). LIBERO evaluation uses 4$\times$ RTX 3090 (CUDA 12.6).

\subsection{Research Questions}\label{sec:rqs}

We organize the experiments around three questions: \textbf{(RQ1)}~How do the four methods compare in task performance as the delay $d$ increases? \textbf{(RQ2)}~How well do training-based methods generalize when evaluated at delays beyond $d_{\max}$? \textbf{(RQ3)}~How do training cost and inference overhead trade off across methods at small (Kinetix) vs.\ VLA (LIBERO) scale?

\section{Kinetix Results}\label{sec:kinetix_results}

\subsection{RQ1: Scaling with Delay}\label{sec:rq1}

\begin{figure}[h]
    \centering
    \includegraphics[width=\textwidth]{\kinetixfig{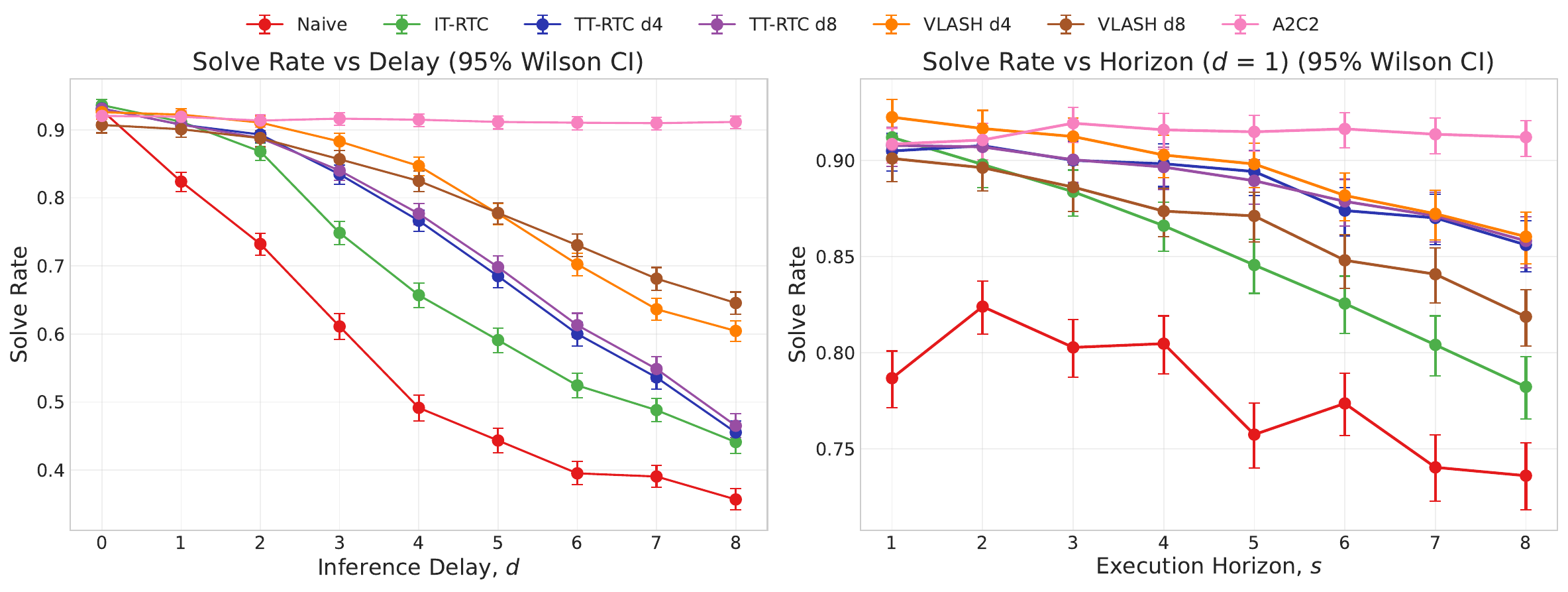}}
    \caption{Kinetix at $H{=}16$ across 10 environments (2048 rollouts each). \emph{Left:} solve rate vs.\ delay $d$. \emph{Right:} solve rate vs.\ horizon $s$ at $d{=}1$. Seven variants: naive, IT-RTC, TT-RTC ($d_{\max} \in \{4, 8\}$), VLASH ($d_{\max} \in \{4, 8\}$), A2C2.}
    \label{fig:kinetix_c16_combined}
\end{figure}

Figure~\ref{fig:kinetix_c16_combined} shows the main $H{=}16$ results.

In the delay sweep (left), A2C2 (pink) achieves the highest absolute solve rate across nearly all delays, remaining above 90\% even at $d{=}8$, while the naive baseline drops below 40\%. TT-RTC with $d_{\max}{=}8$ closely tracks IT-RTC at $d{=}0,1$ and surpasses it for higher delays, confirming that training-time delay simulation can match or exceed inference-time guidance without the VJP overhead. VLASH outperforms the RTC methods at low delays but falls behind A2C2. The horizon sweep at $d{=}1$ (right) shows that all methods except naive remain stable across $s$ with minimal drops in performance ($5-15$\% only), with IT-RTC, TT-RTC, and VLASH ($d_{\max}{=}4$) exhibiting the most consistent curves.

\begin{figure}[h]
    \centering
    \includegraphics[width=\textwidth]{\kinetixfig{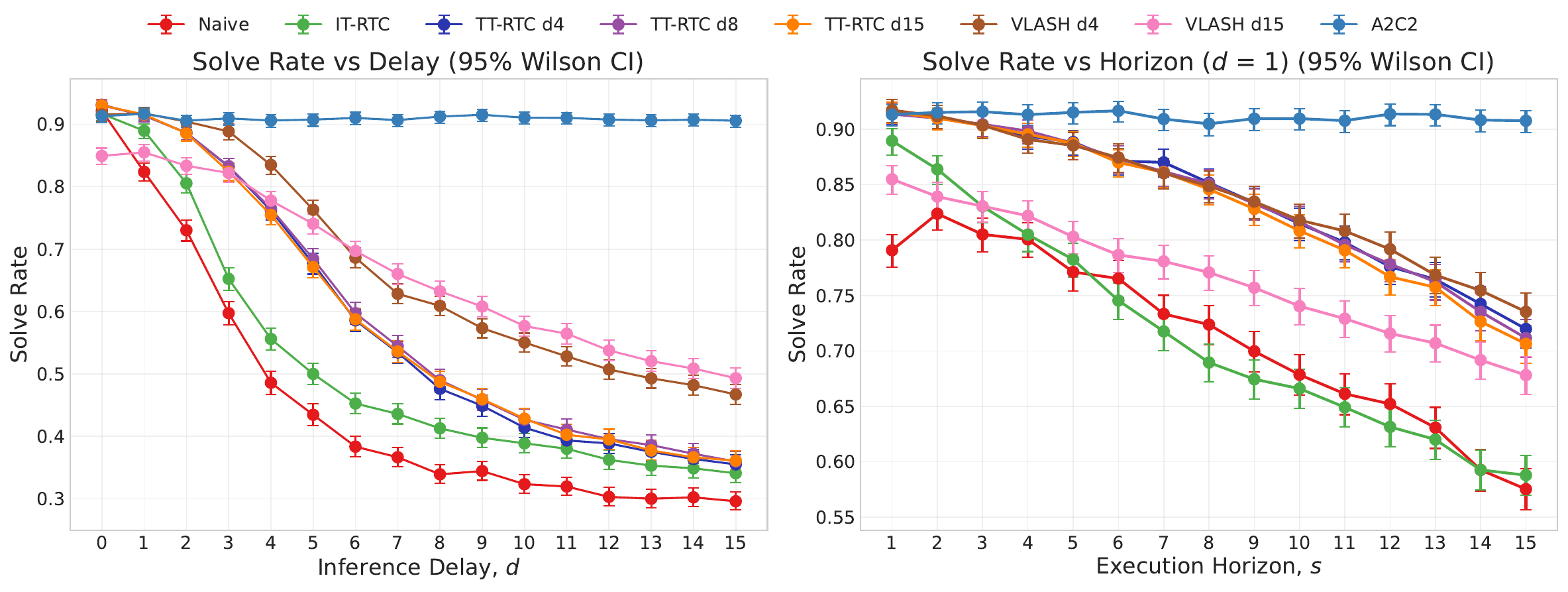}}
    \caption{Kinetix at $H{=}30$, extending the delay range to $d{=}15$. Eight variants: naive, IT-RTC, TT-RTC ($d_{\max} \in \{4,8,15\}$), VLASH ($d_{\max} \in \{4,15\}$), A2C2.}
    \label{fig:kinetix_c30_combined}
\end{figure}

To probe higher delays representative of real-world VLA deployment ($d{=}15$ corresponds to 300\,ms at 50\,Hz; cf.~Table~\ref{tab:delay_mapping}), we repeat the evaluation with $H{=}30$ (Figure~\ref{fig:kinetix_c30_combined}).

The overall ranking persists: A2C2 leads, followed by VLASH ($d_{\max}{=}4$) and the TT-RTC family, with IT-RTC and naive trailing. Two new observations stand out. First, IT-RTC now barely improves over naive at high delays, suggesting that its linear VJP approximation degrades as the prefix-to-postfix ratio grows. Second, training VLASH with a wider range ($d_{\max}{=}15$) helps at $d \geq 6$ but costs nearly 10\% absolute solve rate at low delays compared to $d_{\max}{=}4$. At $d{=}0$ all methods recover comparable performance to the base policy, confirming that the asynchronous mechanisms do not introduce regressions at zero delay. Per-environment breakdowns and $d{=}0$ sanity-check plots are provided in Appendix~\ref{app:per_env}.

\subsection{RQ2: Choosing the Training Delay $d_{\max}$}\label{sec:rq2}

\begin{figure}[h]
    \centering
    \includegraphics[width=\textwidth]{\kinetixfig{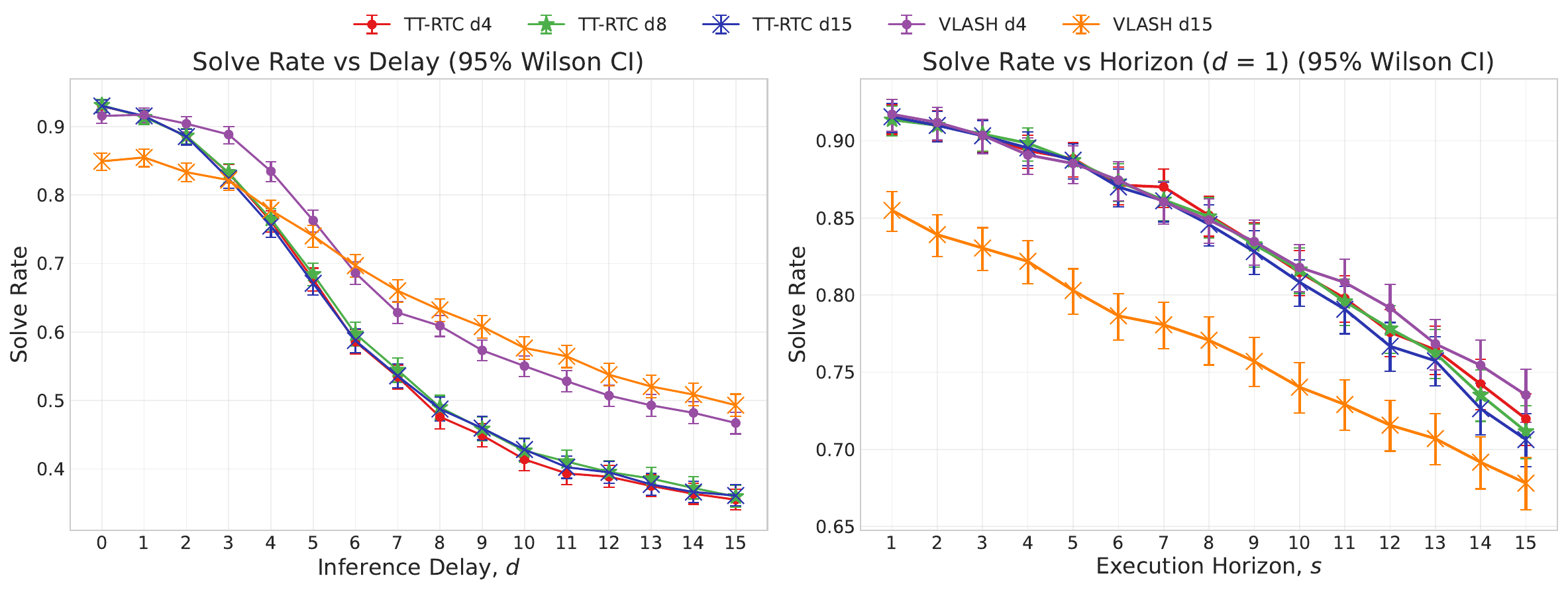}}
    \caption{Effect of $d_{\max}$ on TT-RTC and VLASH at $H{=}30$. \emph{Left:} solve rate vs.\ delay. \emph{Right:} solve rate vs.\ horizon at $d{=}1$. Delays beyond $d_{\max}$ are out-of-distribution for the respective variant.}
    \label{fig:kinetix_dmax_c30}
\end{figure}

Figure~\ref{fig:kinetix_dmax_c30} isolates TT-RTC and VLASH at $H{=}30$ across $d_{\max}$ values.

A clear distinction emerges between the two fine-tuning-based methods: TT-RTC is largely insensitive to $d_{\max}$: the $d_{\max} \in \{4, 8, 15\}$ curves overlap, and even out-of-distribution delays degrade gracefully. VLASH shows the opposite pattern: a wider $d_{\max}$ consistently improves robustness at high delays but measurably hurts low-delay accuracy. We attribute this to the underlying mechanism: VLASH's temporal-offset augmentation directly conditions on the offset, so the training distribution shifts the policy's bias; TT-RTC's prefix conditioning provides a more natural extrapolation. The same trend, less pronounced, appears at $H{=}16$ (figure in Appendix~\ref{app:per_env}). Cross-chunk-size comparisons confirming that the $H{=}16$ and $H{=}30$ ablations are mutually consistent over the shared range $d \in \{0, \ldots, 8\}$ are deferred to Appendix~\ref{app:cross_chunk}.

\paragraph{Practical takeaway.} For VLASH, $d_{\max}$ must be tuned to the expected deployment delay; for TT-RTC, choosing $d_{\max}$ generously seems to be essentially free.

\section{LIBERO Results}\label{sec:libero_results}

We evaluate the same methods on LIBERO with SmolVLA, with $d \in \{0,1,2,4,8,15,20\}$ and $s{=}50$. As noted above, the LIBERO VLASH implementation accesses ground-truth future states; results for VLASH should be read as an \emph{upper bound} on the achievable performance of a deployable variant.

\begin{figure}[h]
    \centering
    \includegraphics[width=0.65\textwidth]{\liberofig{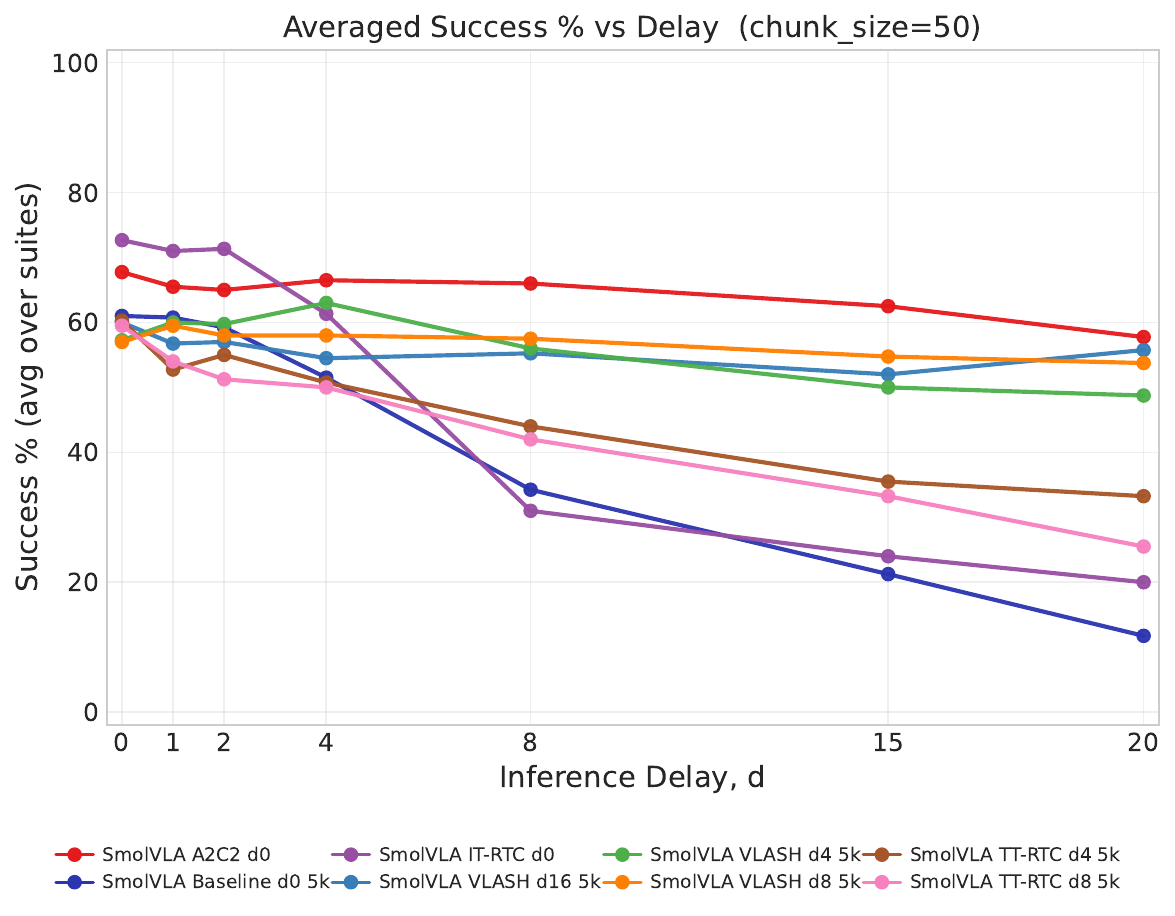}}
    \caption{LIBERO success rate averaged across the four suites (Spatial, Object, Goal, 10-task) vs.\ inference delay $d$, with $s{=}50$. SmolVLA is the base policy; 10 rollouts per task per $(d,s)$.}
    \label{fig:libero_avg_success}
\end{figure}

The results on LIBERO are meaningfully different from Kinetix: at $d{=}0$, IT-RTC achieves the highest success rate ($\sim$75\%), with A2C2 second ($\sim$70\%) and the VLASH and TT-RTC variants clustered between $\sim$57\textendash 61\%. As delay grows, this ranking inverts almost completely. The naive baseline collapses fastest, falling from $\sim$61\% at $d{=}0$ to $\sim$10\textendash 12\% at $d{=}20$, and IT-RTC follows it closely from above\textemdash from $\sim$75\% at $d{=}0$ down to $\sim$20\% at $d{=}20$\textemdash and especially beyond $d{=}8$. TT-RTC also degrades substantially: $d_{\max}{=}8$ falls from $\sim$60\% to $\sim$25\% at $d{=}20$ and $d_{\max}{=}4$ from $\sim$57\% to $\sim$33\%, in marked contrast to TT-RTC's near-flat profile on Kinetix.

In contrast, A2C2 and the wider VLASH variants are remarkably stable. VLASH $d_{\max}{\in}\{8,16\}$ traces nearly horizontal curves, retaining $\sim$55\textendash 56\% through $d{=}20$ (within $\sim$5 points of their $d{=}0$ values); VLASH $d_{\max}{=}4$ instead peaks at $d{=}4$ ($\sim$63\%) and declines more steeply afterwards, replicating the low-delay/high-delay $d_{\max}$ trade-off observed on Kinetix (RQ2). A2C2 takes the lead over every other method from $d{=}4$ onwards and holds a $\sim$10-point margin over the next-best variants throughout $d \in \{4, 8, 15\}$ (mid-to-high 60s vs.\ mid-50s); at $d{=}20$ that margin shrinks as A2C2 begins to slip ($\sim$58\%) while VLASH $d_{\max}{=}16$ ($\sim$56\%) and $d_{\max}{=}8$ ($\sim$55\%) remain nearly flat, with $d_{\max}{=}4$ trailing at $\sim$49\%. The per-suite breakdown (Appendix~\ref{app:libero_per_suite}) shows the same ranking holds across LIBERO-Spatial, -Object, -Goal, and -10, with the largest method gaps on the LIBERO-Spatial and -Object suites. A VLASH ablation over $d_{\max} \in \{0,4,8,16\}$ at multiple training checkpoints (Appendix~\ref{app:libero_vlash_ablation}) further confirms that wider $d_{\max}$ values improve high-delay robustness.

\paragraph{Why does the Kinetix ranking not transfer?} First, SmolVLA action chunks are much longer ($H{=}50$ vs.\ $H{\in}\{16,30\}$ on Kinetix), which inflates the prefix-to-postfix ratio at any given $d$ and may handicap the prefix-conditioned methods (IT-RTC, TT-RTC). Second, LIBERO VLASH benefits from the privileged future-state access discussed in Section~\ref{sec:benchmarks}, which would not be available in real-world deployment; its strong performance at high delay should therefore be read as an upper bound. We did test an analogous variant on Kinetix and found it to track the standard VLASH implementation closely, suggesting the privileged shortcut is a faithful proxy for the deployable mechanism, but this is not conclusive given how different the two environments are. A2C2 also no longer dominates at the very lowest delays as it did on Kinetix\textemdash IT-RTC narrowly beats it at $d \in \{0, 1, 2\}$, after which A2C2 retakes the lead\textemdash potentially because the base SmolVLA's actions are already close to optimal at low delays. This leaves the residual head with a higher chance of injecting noise, as it is trained with positions sampled uniformly within the chunk and is not aware of a specific delay, so it cannot specialize to the small-prefix regime the way IT-RTC does.

\section{Cost Analysis and Inference Overhead}\label{sec:cost}

We compare the methods on three cost axes: per-chunk inference FLOPs, additional training FLOPs, and measured wall-clock latency. Detailed FLOPs derivations and break-even calculations are in Appendix~\ref{app:cost}.

\paragraph{Inference FLOPs.} On Kinetix's compact MLP-Mixer ($F_{\text{fwd}}^K \approx 47.1$\,MFLOPs), IT-RTC's pseudoinverse guidance costs $3 F_{\text{fwd}}^K$ per Euler step (one forward + one VJP backward at $\approx 2\times$ forward~\citep{griewank2008evaluating}) for a $3\times$ overhead. TT-RTC, VLASH, and A2C2 add zero or negligible model FLOPs. On LIBERO, the picture flips: SmolVLA's vision-encoder prefill ($F_{\text{pre}} = 494.6$\,GFLOPs measured via \texttt{torch.profiler}) accounts for 89.8\% of the chunk cost, so IT-RTC's $3\times$ denoising-loop overhead translates to only $1.20\times$ overall ($663.8 / 551.0$\,GFLOPs). A2C2 with $s{=}50$ incurs the largest per-chunk overhead on LIBERO: 50 ResidualTransformer forwards (8.21\,GFLOPs each) bring the chunk cost to $1.75\times$ naive---more expensive than IT-RTC.

\paragraph{Training FLOPs and break-even.} TT-RTC's 8-epoch fine-tune adds $\sim$25\% of base training cost on Kinetix, and 5{,}000 steps of SmolVLA fine-tune ($\sim$1{,}310\,PFLOPs) on LIBERO, with the chosen training hyperparameters in Appendix~\ref{app:hyperparams}. The break-even between TT-RTC's one-time training and IT-RTC's recurring inference overhead is $N^*_{\text{ep}} \approx 2.9 \times 10^6$ episodes on LIBERO. In pure FLOPs terms IT-RTC is therefore quite competitive at VLA scale, but the comparison must also account for sequential latency, since IT-RTC's overhead falls entirely on the (sequential, non-pipelinable) denoising loop.

\paragraph{Measured LIBERO latency.} Table~\ref{tab:latency_libero} reports wall-clock chunk-generation latency on a single LIBERO-Spatial observation on an RTX~3090 (10 warmup + 100 timed iterations). TT-RTC matches the naive baseline almost exactly (402.7 vs.\ 405.2\,ms), IT-RTC is 64.5\,ms slower per chunk (+15.9\%, throughput drops from 2.47 to 2.13\,Hz), and A2C2's residual head is lightweight in isolation (7.27\,ms) so a serial base+residual call sits at +1.8\%.

\begin{table}[h]
\centering
\caption{Measured LIBERO chunk-generation latency on an NVIDIA RTX~3090 for one LIBERO-Spatial observation (SmolVLA, batch size 1). \mbox{``A2C2 residual only''} is reported as a breakdown of the serial path, not as a standalone method.}
\label{tab:latency_libero}
\begin{tabular}{l r r r r}
\toprule
\textbf{Method} & \textbf{Mean (ms)} & \textbf{P95 (ms)} & \textbf{Hz} & \textbf{Ratio} \\
\midrule
Naive               & 405.2 & 407.5 & 2.47  & $1.00\times$ \\
TT-RTC              & 402.7 & 404.9 & 2.48  & $0.99\times$ \\
IT-RTC              & 469.7 & 488.6 & 2.13  & $1.16\times$ \\
A2C2 (serial)       & 412.4 & 417.2 & 2.42  & $1.02\times$ \\
A2C2 residual only  &   7.27 &   7.35 & 137.5 & $0.02\times$ \\
\bottomrule
\end{tabular}
\end{table}

\section{Discussion}\label{sec:discussion}

Across both benchmarks the naive asynchronous baseline is, as expected, the most sensitive to delay. All four mitigation methods recover most of the lost performance, but no single method dominates across all settings, and the cross-benchmark ranking changes more than we initially expected. \textbf{Prefix-conditioned training (TT-RTC)} is the most stable choice on Kinetix: a modest fine-tune, least sensitive to out-of-distribution delays regardless of the training $d_{\max}$, zero inference overhead, and on LIBERO it matches the wall-clock latency of the unmodified base policy. However, on LIBERO its accuracy degrades substantially with delay, eroding much of the appeal of its low cost. \textbf{IT-RTC} is competitive at zero or near-zero delay\textemdash on LIBERO it is the single best method at $d{=}0$\textemdash but its gains shrink as the prefix-to-postfix ratio grows, it triples the sequential denoising-loop latency, and on LIBERO it collapses essentially to the naive baseline beyond $d{=}8$. \textbf{VLASH} exposes a clear low-delay/high-delay trade-off via $d_{\max}$ on both benchmarks, but the $d_{\max}{\in}\{8,16\}$ variants are essentially flat on LIBERO; we re-emphasize that this stability is partly enabled by the privileged future-state access in our LIBERO setup, so a deployable VLASH variant would need to bring its own forward dynamics model. \textbf{A2C2} is the most chunk-size-tolerant method in our experiments: it led on Kinetix, and on LIBERO it is the strongest method at high delay despite the longer chunks. Its only sizeable cost\textemdash accumulating $s$ residual calls per chunk\textemdash is precisely what gives it this robustness, and its design composes naturally with the other three methods, a direction we view as promising and did not exhaustively explore.

\paragraph{Limitations.} Our evaluation treats $d$ as a fixed integer; in real deployments $d$ varies with GPU scheduling, network jitter, and input complexity, conditions under which delay-aware methods (TT-RTC, A2C2) likely have an additional advantage we do not capture. Hyperparameters were taken from the original papers and not individually retuned, as our goal was to get a first understanding of how these recent asynchronous inference approaches compare, rather than developing a new method. Following the VLASH authors' own LIBERO setup, our VLASH results rely on oracle future states to bypass the action/state dimension mismatch that prevents the additive rollforward used in the original VLASH formulation (and possible on Kinetix); a deployable variant would require either a lightweight learned forward model or an environment patch restoring that coupling. Time-to-solve measurements did not prove reliable in simulation, and we leave to future work an evaluation on physical robots, which will be required to draw firm conclusions about energy, time-to-solve, and overall efficiency.

\section{Conclusion}\label{sec:conclusion}

We presented a systematic comparison of four asynchronous action-chunking methods (IT-RTC, TT-RTC, VLASH, A2C2) under controlled conditions, enabled by two unified codebases. On Kinetix across 10 environments and chunk sizes $H \in \{16, 30\}$, A2C2 leads on absolute solve rate, TT-RTC offers the most attractive training/inference trade-off, and TT-RTC overtakes IT-RTC at the highest delays once trained with a wide enough $d_{\max}$. On LIBERO with SmolVLA ($H{=}50$), the ranking shifts: A2C2 is the strongest method at high delay, the wider VLASH variants are also remarkably stable (with the caveat of privileged future-state access), and IT-RTC and TT-RTC degrade substantially as the prefix-to-postfix ratio grows. Training-based methods generalize beyond their training delay distribution when $d_{\max}$ is set appropriately, and prefix-conditioned training (TT-RTC) is markedly more $d_{\max}$-robust than offset-augmented training (VLASH) on Kinetix. In practise: if retraining is infeasible and the expected delay is low, IT-RTC is a sensible drop-in; if fine-tuning is possible and chunks are short, TT-RTC offers a near-free fix; on long-chunk VLA settings, A2C2 is the most reliable choice and naturally composes with the other methods. VLASH remains a strong choice provided an accurate forward state model is available and one can commit to an expected delay range at fine-tune time. All related code is available at: \url{https://github.com/TheAyos/async-vla-inference}.

\begin{ack}
This work was conducted during an exchange semester at ETH Z\"urich and is based on a bachelor's thesis submitted in partial fulfillment of the requirements for a B.Sc. degree at \'Ecole polytechnique.

The author would like to thank Prof.\ Wenqi Jiang for guidance and supervision throughout this project, and the Systems Group at ETH Z\"urich for providing a stimulating research environment and access to computing resources. The author is also grateful to \'Ecole polytechnique for academic and administrative support during this work.

Generative artificial intelligence tools were used to improve the grammar and readability of portions of this manuscript.
\end{ack}

\bibliographystyle{unsrtnat}
\bibliography{neurips_2026_references}

\appendix

\section{Detailed Method Descriptions}\label{app:method_details}

\subsection{IT-RTC: Pseudoinverse Guidance Details}

The soft mask used to blend frozen and free actions is
\begin{equation}
    W_i = c_i \cdot \frac{e^{c_i} - 1}{e - 1}, \qquad c_i = \max(0, 1 - i/d),
\end{equation}
with $W_i = 1$ for fully frozen actions and $W_i = 0$ for fully free actions. During each Euler denoising step $A^{\tau+1/n} = A^\tau + (1/n) v_\theta(A^\tau, o, \tau)$, IT-RTC adds a pseudoinverse-guidance term~\citep{song2023pseudoinverseguided} to the velocity field:
\begin{equation}
    v_{\text{guided}} = v_\theta + \min\!\left(\beta, \tfrac{1-\tau}{\tau \cdot r_\tau^2}\right) \cdot (Y - \hat{A}_t)^{\!\top} \, \mathrm{diag}(W) \cdot \tfrac{\partial \hat{A}_t}{\partial A^\tau},
\end{equation}
where $Y$ is the target prefix, $\hat{A}_t = A^\tau + (1{-}\tau) v_\theta$ is the current denoised estimate, and $\beta$ is a maximum guidance-strength parameter. The Jacobian-vector product $\partial \hat{A}_t / \partial A^\tau$ is computed via \texttt{jax.vjp}, costing one extra backward pass per Euler step (Appendix~\ref{app:cost}).

\subsection{TT-RTC: Architecture Modifications}

TT-RTC trains $\pi_\theta(A_{t+d:H} \mid o_t, A_{t:t+d})$ via three minimal architecture modifications, with no additional parameters. (i)~\emph{Per-token flow timesteps}: the adaLN-Zero conditioning~\citep{dit} is modified to support different flow-matching timesteps for prefix and postfix tokens; prefix actions use $\tau{=}1$ (fully denoised, ground truth) while postfix actions use the sampled training timestep. (ii)~\emph{Non-noisy prefix}: at training time the prefix actions are set to ground truth rather than being noised. (iii)~\emph{Masked loss}: the flow-matching loss is computed only over the postfix. The training-time delay $d$ is sampled from $p(d) \propto \lambda^d$ for $d \in \{0, \ldots, d_{\max}\}$. At inference, TT-RTC is a drop-in replacement for IT-RTC with zero overhead.

\subsection{VLASH: State Rollforward and Temporal-Offset Augmentation}

For position-controlled robots where actions represent target displacements, the future state estimate is
\begin{equation}
    \hat{s}_{t+\Delta} = s_t + \sum_{k=0}^{\Delta-1} a_{t+k}^{\text{prev}},
\end{equation}
where $a_{t+k}^{\text{prev}}$ are the remaining actions from the previous chunk. For velocity- or torque-controlled systems the rollforward is approximate~\citep{vlash}. The training augmentation samples $\delta \in \{0, \ldots, \Delta_{\max}\}$ and trains $\pi_\theta(A_{t+\delta:t+\delta+H-1} \mid o_t, s_{t+\delta})$, keeping the visual observation $o_t$ fixed across $\delta$ so the model learns to rely on the (varying) proprioceptive state for time-sensitive positioning. With block-sparse attention, this can be processed in a single forward pass when state conditioning is routed through a dedicated pathway (e.g.\ adaRMS in $\pi_{0.5}$~\citep{pi05}); when state is embedded in the input token sequence (as in SmolVLA~\citep{smolvla}), each offset requires a separate forward pass.

\subsection{A2C2: Residual Head Details}

The correction head outputs
\begin{equation}
    a_{t+k}^{\text{exec}} = a_{t+k}^{\text{base}} + \Delta a_{t+k}, \qquad \Delta a_{t+k} = \pi_{\text{A2C2}}(o_{t+k}, a_{t+k}^{\text{base}}, \tau_k, z_{t+k}, l),
\end{equation}
where $\tau_k = (\sin(2\pi k/H), \cos(2\pi k/H))$ encodes time within the chunk, $z_{t+k}$ is base-policy hidden state, and $l$ is the language instruction. Training is on pre-collected base-policy rollouts with an $L_2$ residual loss; offsets $k$ are sampled uniformly so the head sees varying staleness. The base policy is frozen and so A2C2 is composable with the other three methods.

\section{Hyperparameters}\label{app:hyperparams}

\subsection{Kinetix}

\paragraph{Worked example: base flow policy.} The base imitation-learning flow policy is shared by Naive, IT-RTC, and A2C2, and is the starting point for TT-RTC fine-tuning. Architecture: a 4-layer MLP-Mixer with channel dimension 256, channel hidden dimension 512, token hidden dimension 64, and adaLN-Zero conditioning supporting both scalar and per-token flow timesteps. Training: imitation learning with flow-matching loss for 32 epochs, batch size 512, AdamW with constant learning rate $3\times10^{-4}$ (linear warmup over 1000 steps), weight decay $10^{-2}$, and gradient clipping to norm 10.0; 5 Euler denoising steps at inference. The full pipeline is re-run for each chunk size $H \in \{16, 30\}$.

\paragraph{TT-RTC ($d_{\max} \in \{5, 9, 16\}$).} 8-epoch fine-tune from the 24-epoch base checkpoint, with the same optimizer, batch size, and learning-rate schedule; the only change is that \texttt{simulated\_delay} is set to the desired $d_{\max}$ to enable per-token prefix conditioning and the masked postfix loss.

\paragraph{VLASH ($d_{\max} \in \{5, 9, 16\}$).} Trained from scratch for 32 epochs with the same optimizer and batch size as the base; \texttt{async\_interval} is set to $d_{\max}$ to enable per-batch random observation/action offsets.

\paragraph{A2C2.} A 3-layer MLP residual head ($\text{obs} + \text{base action} + (\sin, \cos)\,\tau \to 256 \to 512 \to \text{action}$) is trained on top of the frozen base flow policy for 16 epochs at batch size 512, with AdamW learning rate $10^{-4}$, weight decay $10^{-3}$, 500 warmup steps, and gradient clipping 10.0.

\paragraph{IT-RTC.} Inference-only over the base flow policy; uses guidance strength $\beta{=}5.0$ and an exponential prefix-attention schedule.

\subsection{LIBERO}\label{app:libero_hyperparams}

\paragraph{Worked example: base $d{=}0$ SmolVLA fine-tune.} All four LIBERO methods derive from a single SmolVLA fine-tune of the publicly released \texttt{HuggingFaceVLA/smolvla\_libero} checkpoint on the combined Spatial+Object+Goal+10 training data; full configuration is reported in Table~\ref{tab:libero_base_hparams}. Only the action expert is trained (the VLM is frozen via \texttt{num\_vlm\_layers}$=0$ and \texttt{load\_vlm\_weights}$=$\texttt{false}, and the action expert is initialized from the smolvla\_libero weights with \texttt{expert\_width\_multiplier}$=0.5$).

\begin{table}[H]
\centering
\caption{LIBERO base $d{=}0$ SmolVLA fine-tune hyperparameters. The VLASH and TT-RTC variants share this configuration and differ only in the fields called out below; A2C2 uses a separate optimizer and architecture (Table~\ref{tab:libero_a2c2_hparams}).}
\label{tab:libero_base_hparams}
\begin{tabular}{lc}
\toprule
\textbf{Field} & \textbf{Value} \\
\midrule
Initial checkpoint                        & \texttt{HuggingFaceVLA/smolvla\_libero} \\
Components fine-tuned                     & action expert only (VLM frozen) \\
Action chunk size $H$                     & 50 \\
Training dataset                          & \texttt{theayos/libero\_image} (Spatial+Object+Goal+10) \\
Optimizer                                 & AdamW \\
Peak / final learning rate                & $4\times10^{-4}$ / $2.5\times10^{-6}$ \\
$\beta_1, \beta_2$                        & 0.9, 0.95 \\
Weight decay                              & $10^{-10}$ \\
Gradient-clip norm                        & 10.0 \\
LR schedule                               & cosine decay with warmup \\
Warmup / decay steps                      & 1{,}000 / 5{,}000 \\
Total training steps                      & 5{,}000 \\
Batch size (per-GPU $\times$ \#GPUs)      & $64 \times 4 = 256$ \\
Mixed precision                           & enabled (Accelerate) \\
\texttt{max\_delay\_steps} (delay aug.)   & 0 \\
Seed                                      & 42 \\
\bottomrule
\end{tabular}
\end{table}

\paragraph{VLASH ($d_{\max} \in \{4, 8, 16\}$).} Configuration identical to the base $d{=}0$ run, except \texttt{max\_delay\_steps} is set to the target $d_{\max}$ to activate VLASH's temporal-offset augmentation. Each $d_{\max}$ variant is fine-tuned independently from \texttt{HuggingFaceVLA/smolvla\_libero} (not from the base $d{=}0$ checkpoint) for 5{,}000 steps.

\paragraph{TT-RTC ($d_{\max} \in \{5, 9\}$).} Fine-tuned from the base $d{=}0$ checkpoint at step 5{,}000 for an additional 5{,}000 steps with \texttt{max\_delay\_steps}$=d_{\max}$ to enable prefix sampling. Per-GPU batch is 128 on 2 GPUs (effective $128 \times 2 = 256$); optimizer and schedule are otherwise unchanged. The $d_{\max}{=}17$ variant is trained fresh from the public smolvla\_libero checkpoint at $128 \times 4 = 512$ on 4 GPUs to fit the longer prefix.

\paragraph{A2C2 (residual transformer).} Trained on offline $(\,\text{observation}, \text{base action}, \text{ground-truth action}\,)$ triplets collected by rolling out the frozen base $d{=}0$ policy on the LIBERO training data. Architecture and optimizer differ from the SmolVLA fine-tunes; values are summarized in Table~\ref{tab:libero_a2c2_hparams}.

\begin{table}[H]
\centering
\caption{A2C2 residual transformer training on LIBERO.}
\label{tab:libero_a2c2_hparams}
\begin{tabular}{lc}
\toprule
\textbf{Field} & \textbf{Value} \\
\midrule
Base policy                       & frozen base $d{=}0$ SmolVLA (step 5{,}000) \\
Residual model parameters         & 32.0\,M total / 20.8\,M trainable \\
Vision backbone                   & ResNet-18 (frozen, ImageNet weights) \\
$d_{\text{model}}$                & 512 \\
Encoder layers / heads            & 6 / 8 \\
Feed-forward dim                  & 2{,}048 \\
Dropout                           & 0.1 \\
Time-in-chunk feature             & $(\sin, \cos)(2\pi k / H)$ \\
Optimizer                         & AdamW \\
Learning rate / weight decay      & $10^{-5}$ / $10^{-5}$ \\
Mixed precision                   & enabled (Accelerate) \\
Effective batch / \#GPUs          & 64 / 1 \\
Total training steps              & 200{,}000 \\
\bottomrule
\end{tabular}
\end{table}

\paragraph{IT-RTC.} Inference-only over the base $d{=}0$ checkpoint; uses execution horizon $s_{\min}{=}10$, guidance strength $\beta{=}5.0$, and an exponential prefix-attention schedule (\texttt{EXP}).

\section{FLOPs and Break-Even Analysis}\label{app:cost}

\subsection{Kinetix}

We analytically compute FLOPs for the MLP-Mixer policy using $2mn$ FLOPs per matmul of dimensions $m \times n$ per token. With $o = 679$ and representative $a = 6$:

\begin{table}[H]
\centering
\caption{Analytically computed FLOPs for Kinetix models. $C$ denotes chunk size, $N_s = 5$ Euler steps. The flow policy has an input projection, 4 MLPMixerBlocks (token + channel mixing + 2 AdaLN projections), a time MLP, and an output projection. The residual MLP is a 3-layer feedforward network.}
\label{tab:flops_kinetix}
\begin{tabular}{l r r}
\toprule
\textbf{Component} & $C{=}16$ & $C{=}30$ \\
\midrule
\multicolumn{3}{l}{\textit{Flow policy (per denoising step)}} \\
\quad Input projection            & 5.60\,M  & 10.50\,M \\
\quad Time MLP (scalar)           & 0.26\,M  & 0.26\,M  \\
\quad 4$\times$ MLPMixerBlock     & 40.89\,M & 73.92\,M \\
\quad Final projection            & 0.31\,M  & 0.35\,M  \\
\midrule
$F_{\text{fwd}}^K$ (1 step)       & \textbf{47.1\,M} & \textbf{85.0\,M} \\
$5 F_{\text{fwd}}^K$ (1 chunk)    & 235\,M  & 425\,M \\
\midrule
\multicolumn{3}{l}{\textit{Residual MLP (per action step)}} \\
$F_{\text{res}}^K$                & \multicolumn{2}{c}{\textbf{0.62\,M}} \\
\bottomrule
\end{tabular}
\end{table}

\begin{table}[H]
\centering
\caption{Per-chunk inference FLOPs on Kinetix ($C{=}16$, $N_s{=}5$, execution horizon $s$). Ratio is relative to naive.}
\label{tab:inference_kinetix}
\begin{tabular}{l l r r}
\toprule
\textbf{Method} & \textbf{Per-chunk FLOPs} & \textbf{MFLOPs} & \textbf{Ratio} \\
\midrule
Naive  & $N_s \cdot F_{\text{fwd}}^K$                                   & 235          & $1.0\times$       \\
IT-RTC & $3 N_s \cdot F_{\text{fwd}}^K$                                 & 707          & $3.0\times$       \\
TT-RTC & $N_s \cdot F_{\text{fwd}}^K$                                   & 235          & $1.0\times$       \\
VLASH  & $N_s \cdot F_{\text{fwd}}^K$ (+ env sim)                        & 235          & $1.0\times$       \\
A2C2   & $N_s \cdot F_{\text{fwd}}^K + s \cdot F_{\text{res}}^K$         & $235 + 0.62s$ & $\approx 1.02\times$ \\
\bottomrule
\end{tabular}
\end{table}

\begin{table}[H]
\centering
\caption{Additional training cost for each method on Kinetix beyond the shared 32-epoch base flow policy ($\approx 54.2$~PFLOPs).}
\label{tab:training_kinetix}
\begin{tabular}{l c r r}
\toprule
\textbf{Method} & \textbf{Config} & \textbf{Additional cost} & \textbf{\% of base} \\
\midrule
IT-RTC  & Training-free                       & 0                          & 0\%   \\
TT-RTC  & 8 epochs FT, $B{=}512$              & $\sim$13.6\,PFLOPs         & 25\%  \\
VLASH   & 32 epochs retrain, $B{=}512$        & $\sim$54.2\,PFLOPs         & 100\% \\
A2C2    & 16 epochs, $B{=}512$                & $\sim$27.4\,PFLOPs$^\dagger$ & 51\%  \\
\bottomrule
\end{tabular}
\smallskip
\footnotesize{$^\dagger$ A2C2 training requires a frozen base-policy forward per step; the residual head cost is negligible ($\approx 1.3\%$ of base forward).}
\end{table}

\subsection{LIBERO}

\begin{table}[H]
\centering
\caption{Measured FLOPs for the LIBERO models (SmolVLA with \texttt{expert\_width\_multiplier}$= 0.5$, $H{=}50$, $N_s{=}10$). Values obtained via \texttt{torch.profiler} on an NVIDIA H100 NVL.}
\label{tab:flops_libero}
\begin{tabular}{l r}
\toprule
\textbf{Component} & \textbf{FLOPs} \\
\midrule
SmolVLA prefill ($F_{\text{pre}}$)                       & 494.6\,GFLOPs \\
SmolVLA denoising step ($F_{\text{ds}}$)                 & 5.64\,GFLOPs  \\
SmolVLA full chunk ($F_{\text{pre}} + 10 F_{\text{ds}}$) & 551.0\,GFLOPs \\
\midrule
ResidualTransformer forward ($F_{\text{res}}^L$)         & 8.21\,GFLOPs  \\
\midrule
SmolVLA total parameters                                  & 604.9\,M \\
ResidualTransformer parameters (total / trainable)        & 32.0\,M / 20.8\,M \\
\bottomrule
\end{tabular}
\end{table}

\begin{table}[H]
\centering
\caption{Per-chunk inference FLOPs on LIBERO. The shared, dominating prefill makes IT-RTC's overhead modest overall but A2C2's per-step head accumulates over $s$.}
\label{tab:inference_libero}
\begin{tabular}{l l r r}
\toprule
\textbf{Method} & \textbf{Per-chunk FLOPs} & \textbf{GFLOPs} & \textbf{Ratio} \\
\midrule
Naive          & $F_{\text{pre}} + 10 F_{\text{ds}}$                                & 551.0 & $1.00\times$ \\
IT-RTC         & $F_{\text{pre}} + 30 F_{\text{ds}}$                                & 663.8 & $1.20\times$ \\
TT-RTC         & $F_{\text{pre}} + 10 F_{\text{ds}}$                                & 551.0 & $1.00\times$ \\
VLASH          & $F_{\text{pre}} + 10 F_{\text{ds}}$ (+ state lookup)               & 551.0 & $1.00\times$ \\
A2C2 ($s{=}50$)& $F_{\text{pre}} + 10 F_{\text{ds}} + 50 F_{\text{res}}^L$          & 961.7 & $1.75\times$ \\
\bottomrule
\end{tabular}
\end{table}

\begin{table}[H]
\centering
\caption{LIBERO training cost. $B_{\text{eff}}$ is the effective batch size. Per-sample cost: $F_{\text{pre}} + 3 F_{\text{ds}} = 511.5$\,GFLOPs/sample for SmolVLA fine-tuning (frozen vision encoder forward + expert forward + backward); $3 F_{\text{res}}^L$ for A2C2.}
\label{tab:training_libero}
\begin{tabular}{l r r r r}
\toprule
\textbf{Method} & \textbf{Steps} & $B_{\text{eff}}$ & \textbf{GPUs} & \textbf{Est.\ total cost} \\
\midrule
Baseline ($d_{\max}{=}0$)   & 5{,}000   & 256 & 4 & $\sim$654\,PFLOPs   \\
TT-RTC ($d_{\max}{=}8$)     & 5{,}000   & 512 & 4 & $\sim$1{,}310\,PFLOPs \\
VLASH ($d_{\max}{=}8$)      & 5{,}000   & 256 & 4 & $\sim$654\,PFLOPs   \\
A2C2 (ResTransformer)       & 200{,}000 & 64  & 1 & $\sim$315\,PFLOPs   \\
\bottomrule
\end{tabular}
\end{table}

\subsection{Break-Even: TT-RTC vs.\ IT-RTC}\label{app:breakeven}

We define
\begin{equation}\label{eq:breakeven}
    N^* = \frac{C_{\text{train}}^{\text{TT-RTC}}}{\Delta C_{\text{chunk}}}, \qquad
    \Delta C_{\text{chunk}} = C_{\text{chunk}}^{\text{IT-RTC}} - C_{\text{chunk}}^{\text{TT-RTC}}.
\end{equation}
Converting to episodes: $N^*_{\text{ep}} = N^* \cdot s / L_{\text{ep}}$.

\paragraph{Kinetix ($C{=}16$).} $C_{\text{train}}^{\text{TT-RTC}} = 8 \times 23{,}437 \times 512 \times 3 F_{\text{fwd}}^K \approx 2.88 \times 10^8 F_{\text{fwd}}^K$ and $\Delta C_{\text{chunk}} = 10 F_{\text{fwd}}^K$, giving $N^* \approx 2.88 \times 10^7$ chunks. With $L_{\text{ep}} \approx 80$ and $s{=}8$ (10 chunks/episode), $N^*_{\text{ep}} \approx 2.88 \times 10^6$ episodes. On 4$\times$H100 NVL, both TT-RTC fine-tuning and millions of evaluation episodes complete in minutes; the choice between IT-RTC and TT-RTC at this scale is performance-driven, not compute-driven.

\paragraph{LIBERO (SmolVLA).} $C_{\text{train}}^{\text{TT-RTC}} \approx 1{,}310$\,PFLOPs and $\Delta C_{\text{chunk}} = 20 F_{\text{ds}} = 112.8$\,GFLOPs, giving $N^* \approx 1.16 \times 10^7$ chunks. With $L_{\text{ep}} \approx 200$ and $s{=}50$ (4 chunks/episode), $N^*_{\text{ep}} \approx 2.9 \times 10^6$ episodes. The break-even is large because training cost is dominated by the frozen vision-encoder forward ($F_{\text{pre}} \gg F_{\text{ds}}$) while the inference overhead acts only on $F_{\text{ds}}$. In purely FLOPs terms IT-RTC is competitive, but its $3\times$ overhead falls entirely on the (sequential, non-pipelinable) denoising loop, whereas the prefill can in principle be pipelined with action execution---so TT-RTC remains preferable for latency-sensitive deployment.

\section{Per-Environment Kinetix Breakdowns}\label{app:per_env}

\begin{figure}[H]
    \centering
    \includegraphics[width=0.9\textwidth]{\kinetixfig{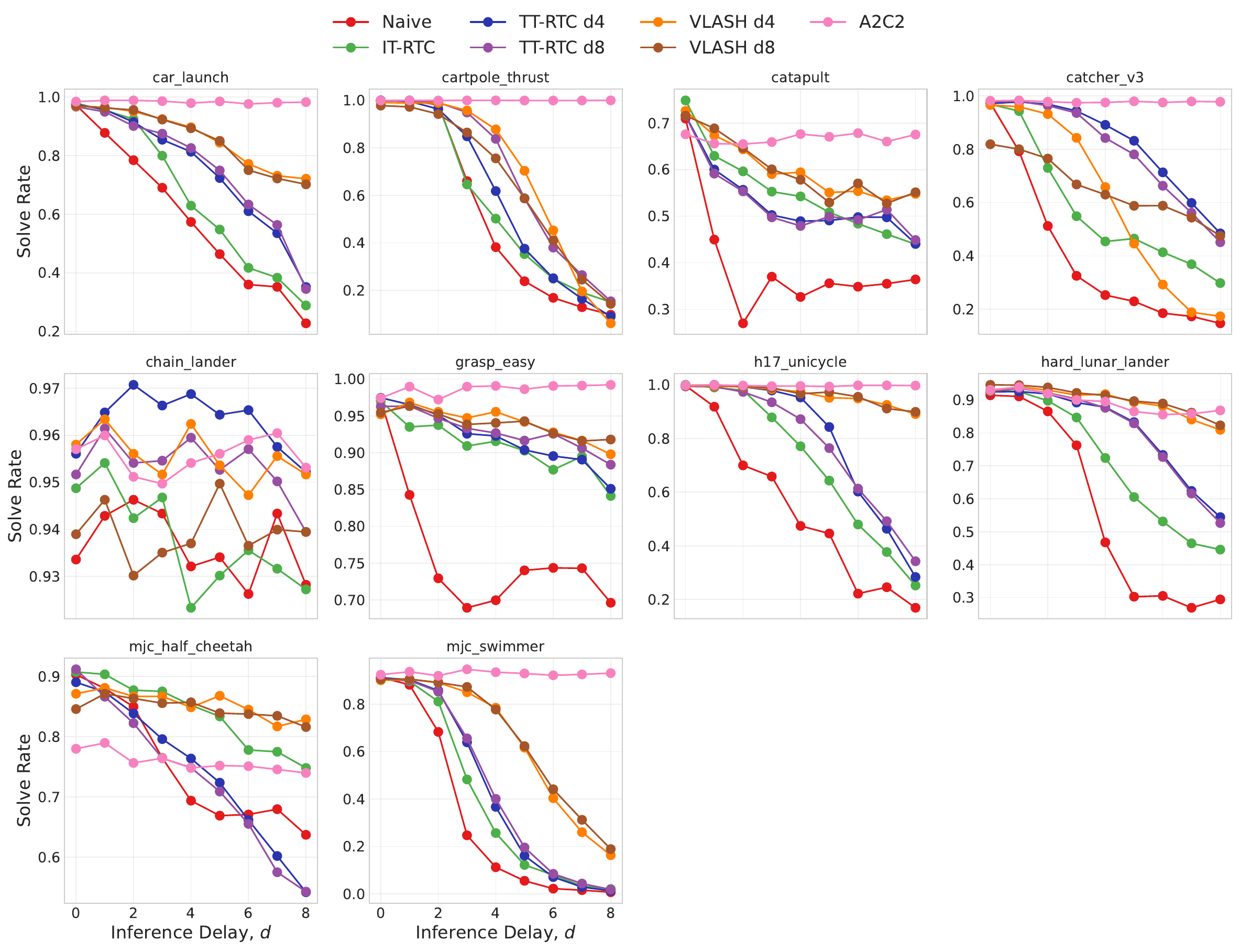}}
    \caption{Per-environment solve rate vs.\ delay at $H{=}16$ (best execution horizon per delay).}
    \label{fig:kinetix_c16_per_env}
\end{figure}

\begin{figure}[H]
    \centering
    \includegraphics[width=0.9\textwidth]{\kinetixfig{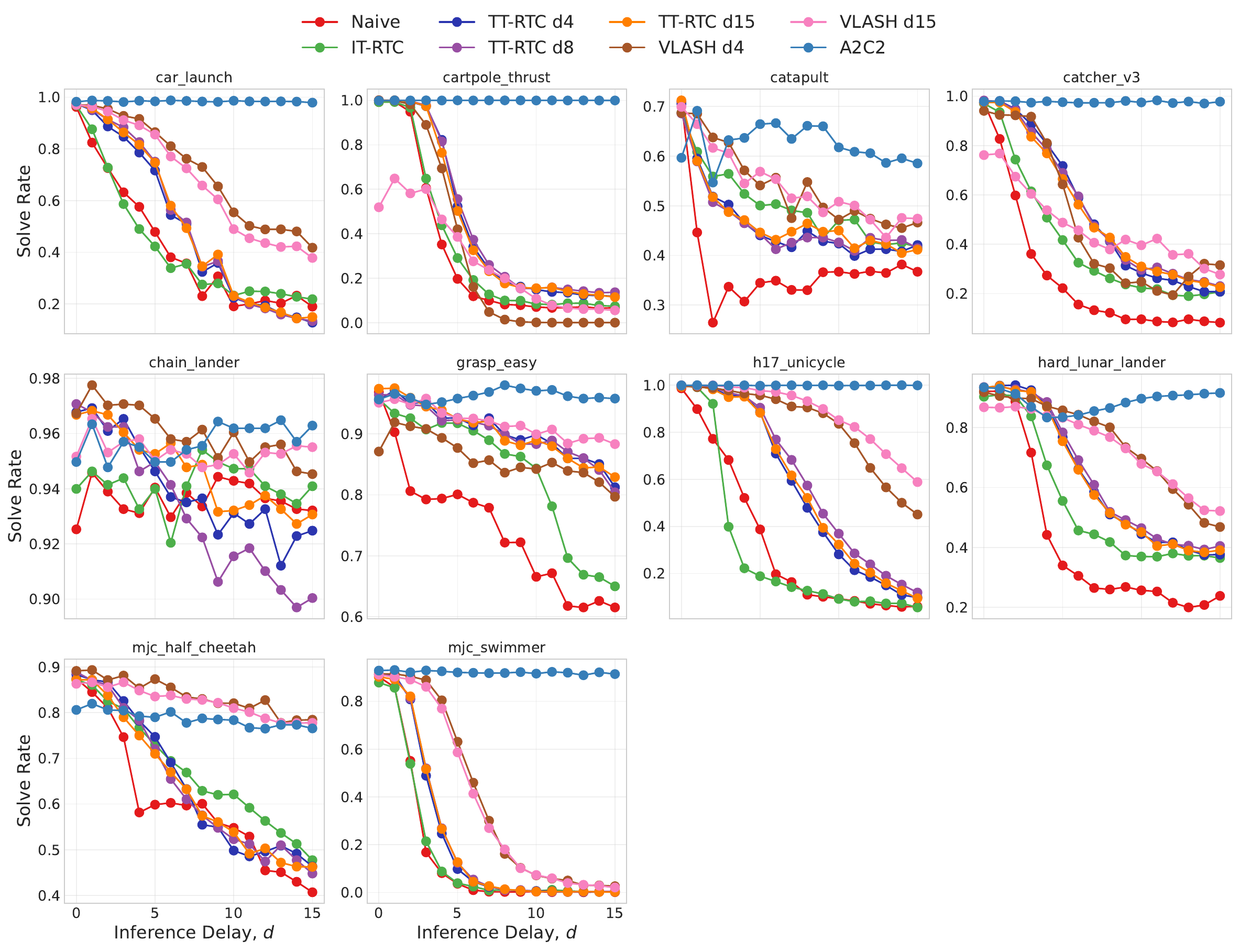}}
    \caption{Per-environment solve rate vs.\ delay at $H{=}30$ (best execution horizon per delay).}
    \label{fig:kinetix_c30_per_env}
\end{figure}

\begin{figure}[H]
    \centering
    \includegraphics[width=\textwidth]{\kinetixfig{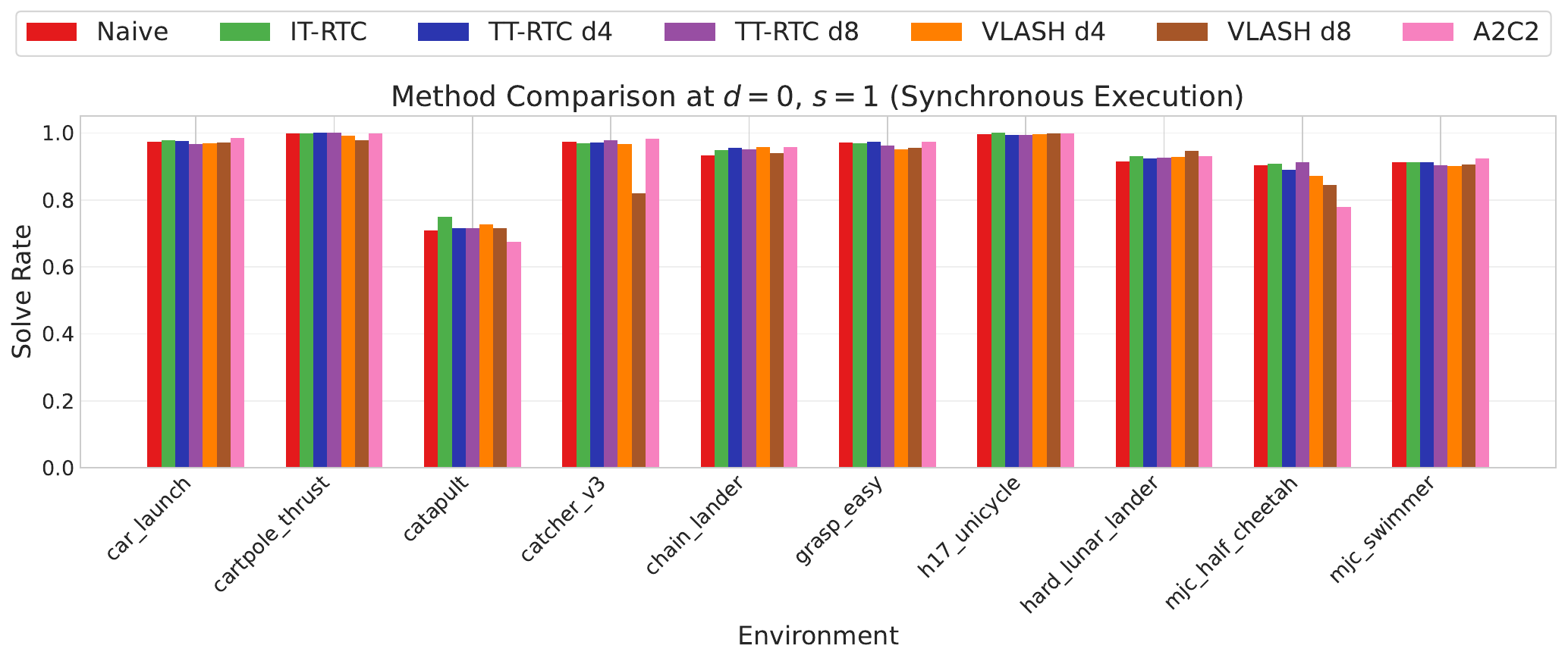}}\\[0.5cm]
    \includegraphics[width=\textwidth]{\kinetixfig{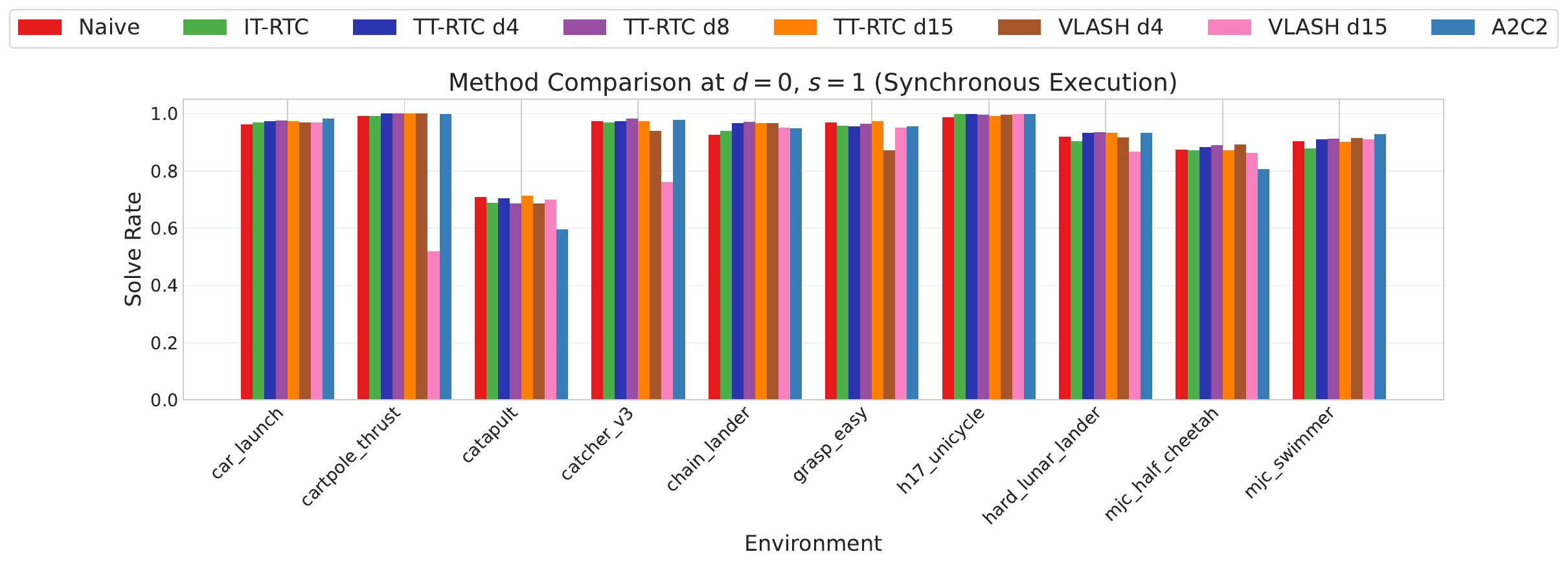}}
    \caption{Per-environment solve rate at $d{=}0$, $s{=}1$ (synchronous execution) for $H{=}16$ (top) and $H{=}30$ (bottom). All variants should ideally match the base policy.}
    \label{fig:kinetix_baseline}
\end{figure}

\begin{figure}[H]
    \centering
    \includegraphics[width=\textwidth]{\kinetixfig{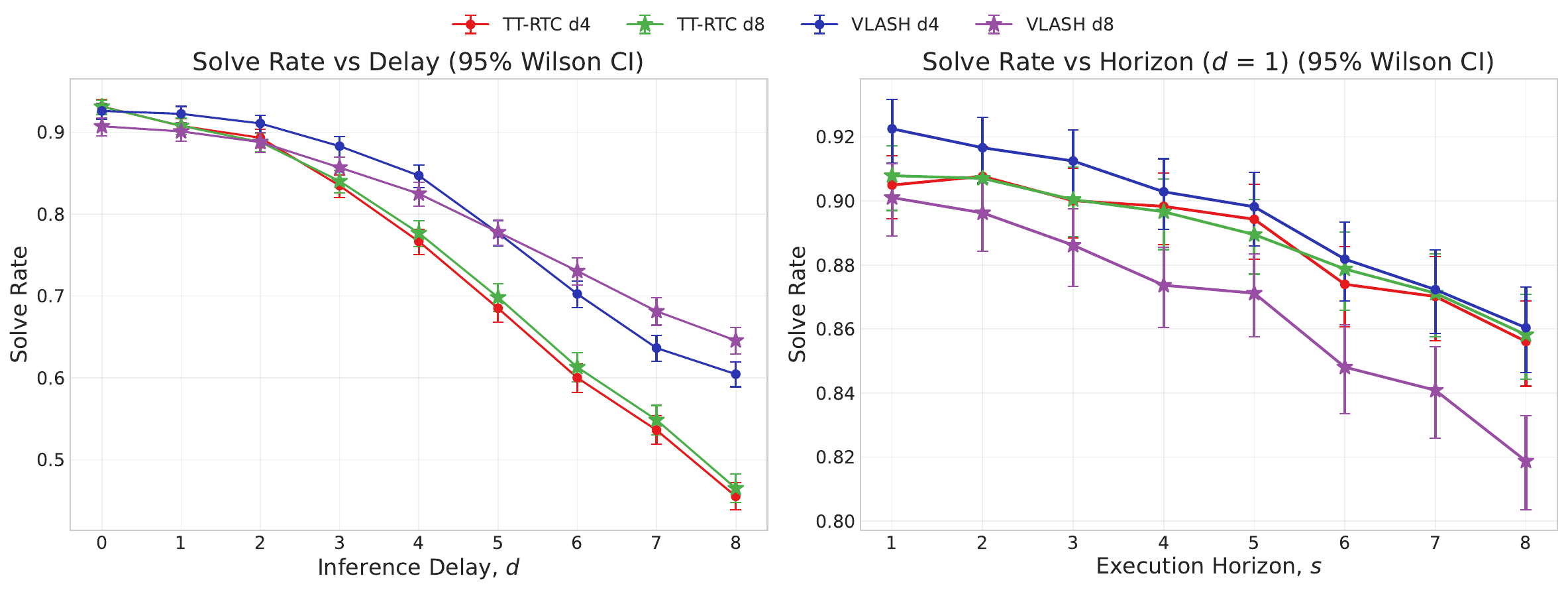}}
    \caption{Effect of $d_{\max}$ on TT-RTC and VLASH at $H{=}16$ (companion to Figure~\ref{fig:kinetix_dmax_c30}).}
    \label{fig:kinetix_dmax_c16}
\end{figure}

\section{Cross-Chunk-Size Kinetix Comparisons}\label{app:cross_chunk}

These plots verify that the $H{=}16$ and $H{=}30$ training-delay ablations are mutually consistent over the shared range $d \in \{0, \ldots, 8\}$. For naive, IT-RTC, and A2C2, $H{=}16$ consistently beats $H{=}30$ in absolute solve rate, confirming that longer chunks impose an intrinsic prediction difficulty even before delay-related degradation; IT-RTC's curves are nearly parallel across chunk sizes, indicating that its VJP guidance scales well with chunk length. TT-RTC's variants are consistent across chunk sizes when $d_{\max}$ is set appropriately. VLASH $H{=}16$, $d_{\max}{=}8$ slightly outperforms VLASH $H{=}30$, $d_{\max}{=}15$ despite the latter's wider training range, suggesting that VLASH's state-rollforward accuracy is more chunk-size-sensitive than TT-RTC's prefix conditioning.

\begin{figure}[H]
    \centering
    \includegraphics[width=\textwidth]{\kinetixfig{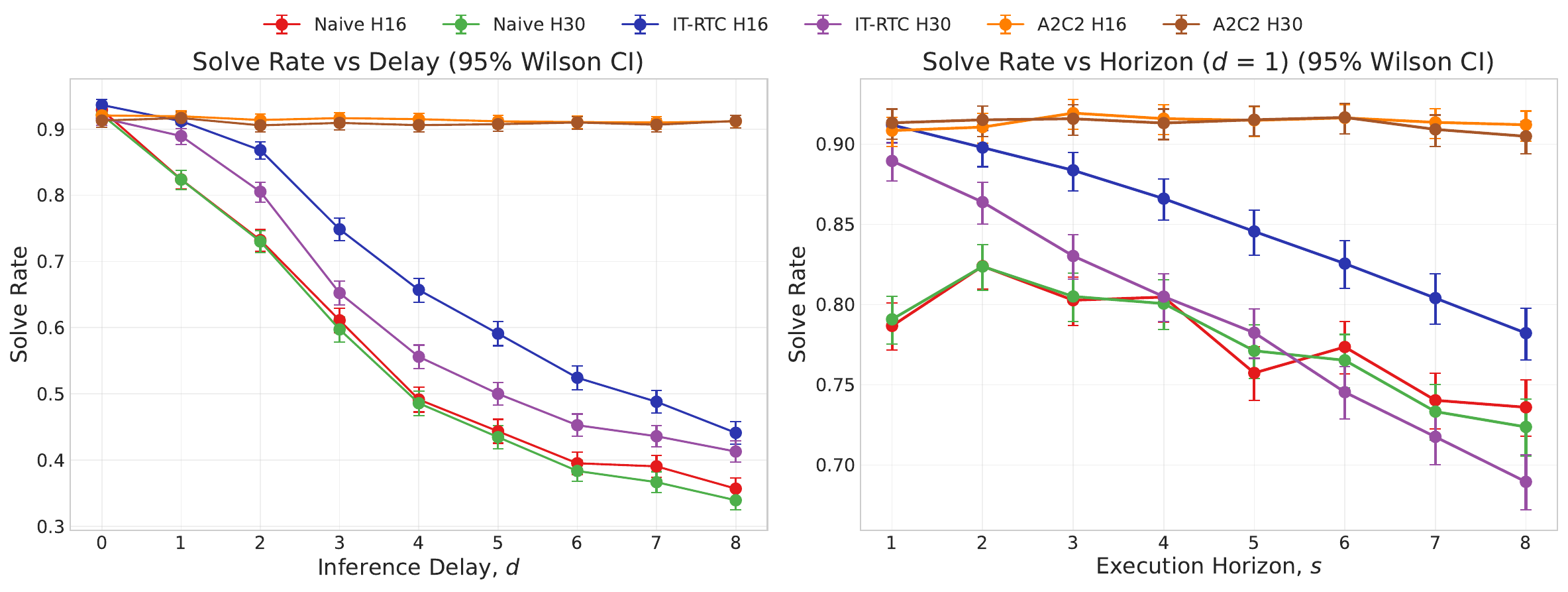}}
    \caption{Cross-chunk comparison for training-free methods (naive, IT-RTC) and A2C2: $H{=}16$ vs.\ $H{=}30$ over $d \in \{0,\ldots,8\}$.}
    \label{fig:kinetix_cross_nir}
\end{figure}

\begin{figure}[H]
    \centering
    \includegraphics[width=\textwidth]{\kinetixfig{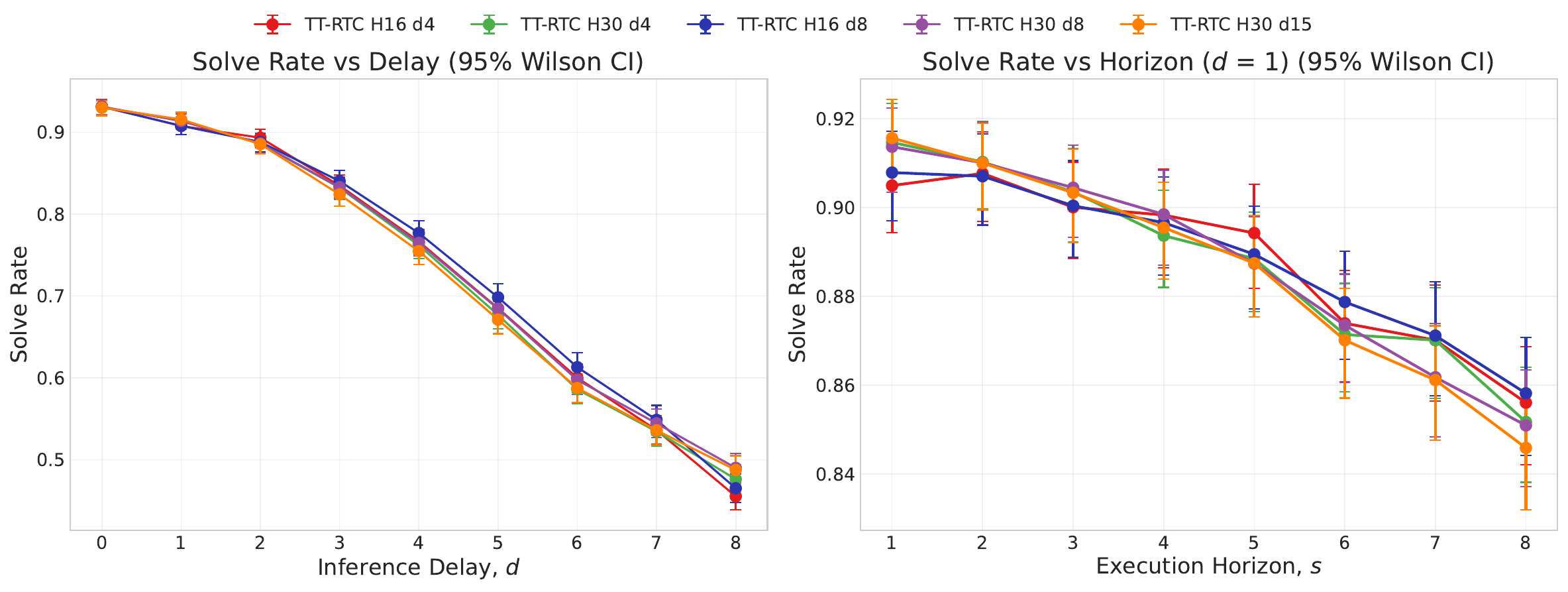}}
    \caption{Cross-chunk comparison for TT-RTC: $H{=}16$ and $H{=}30$ variants across $d_{\max}$ over the shared range.}
    \label{fig:kinetix_cross_ttrtc}
\end{figure}

\begin{figure}[H]
    \centering
    \includegraphics[width=\textwidth]{\kinetixfig{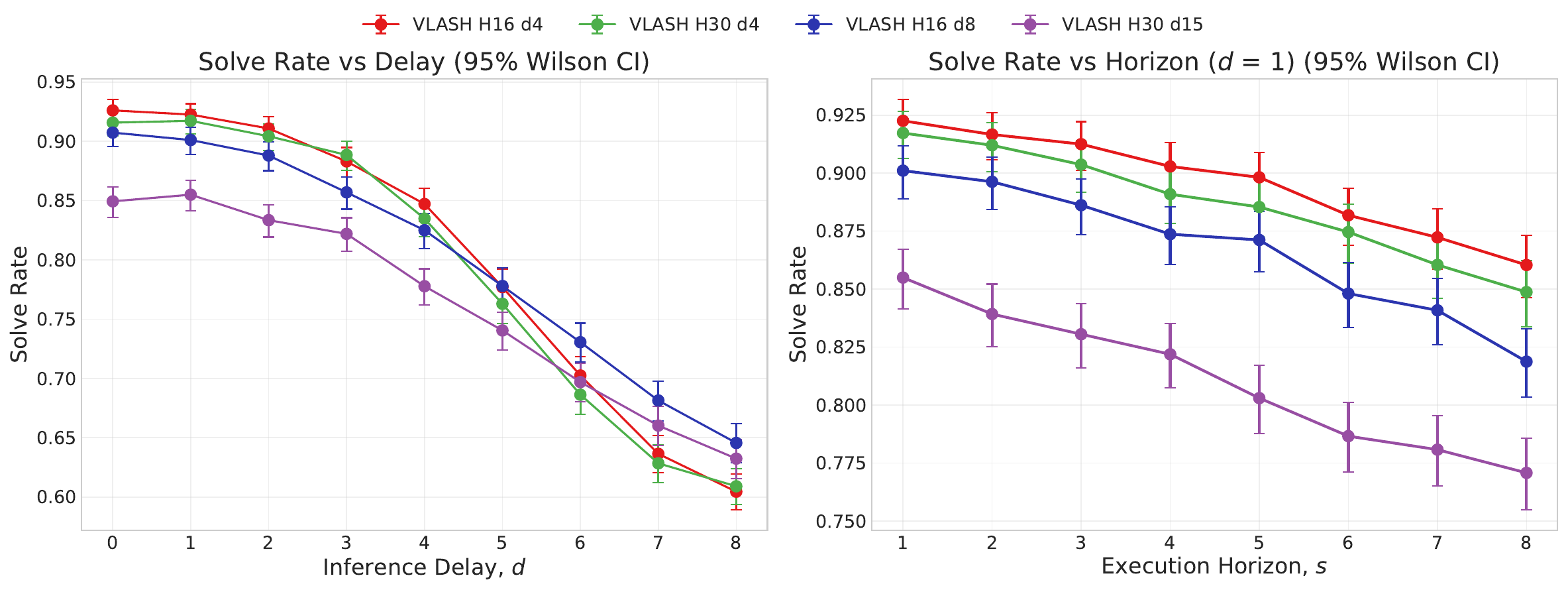}}
    \caption{Cross-chunk comparison for VLASH: $H{=}16$ and $H{=}30$ variants across $d_{\max}$ over the shared range.}
    \label{fig:kinetix_cross_vlash}
\end{figure}

\section{LIBERO: Episode Length and Per-Suite Results}\label{app:libero_per_suite}

\begin{figure}[H]
    \centering
    \includegraphics[width=0.75\textwidth]{\liberofig{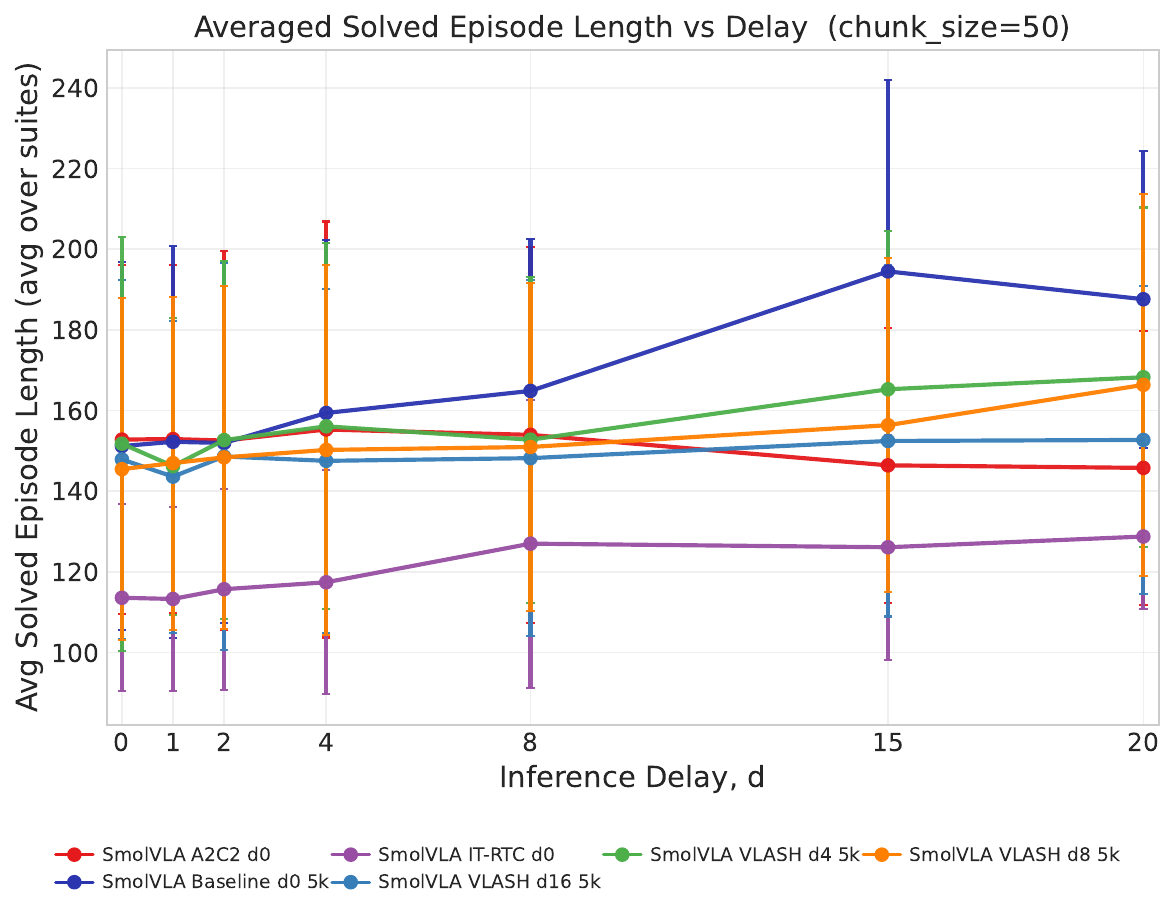}}
    \caption{Average solved-episode length (successful episodes only) vs.\ inference delay $d$, averaged across the four LIBERO suites. Error bars: $\pm 1$ standard deviation pooled across suites.}
    \label{fig:libero_avg_ep_length}
\end{figure}

\begin{figure}[H]
    \centering
    \includegraphics[width=0.95\textwidth]{\liberofig{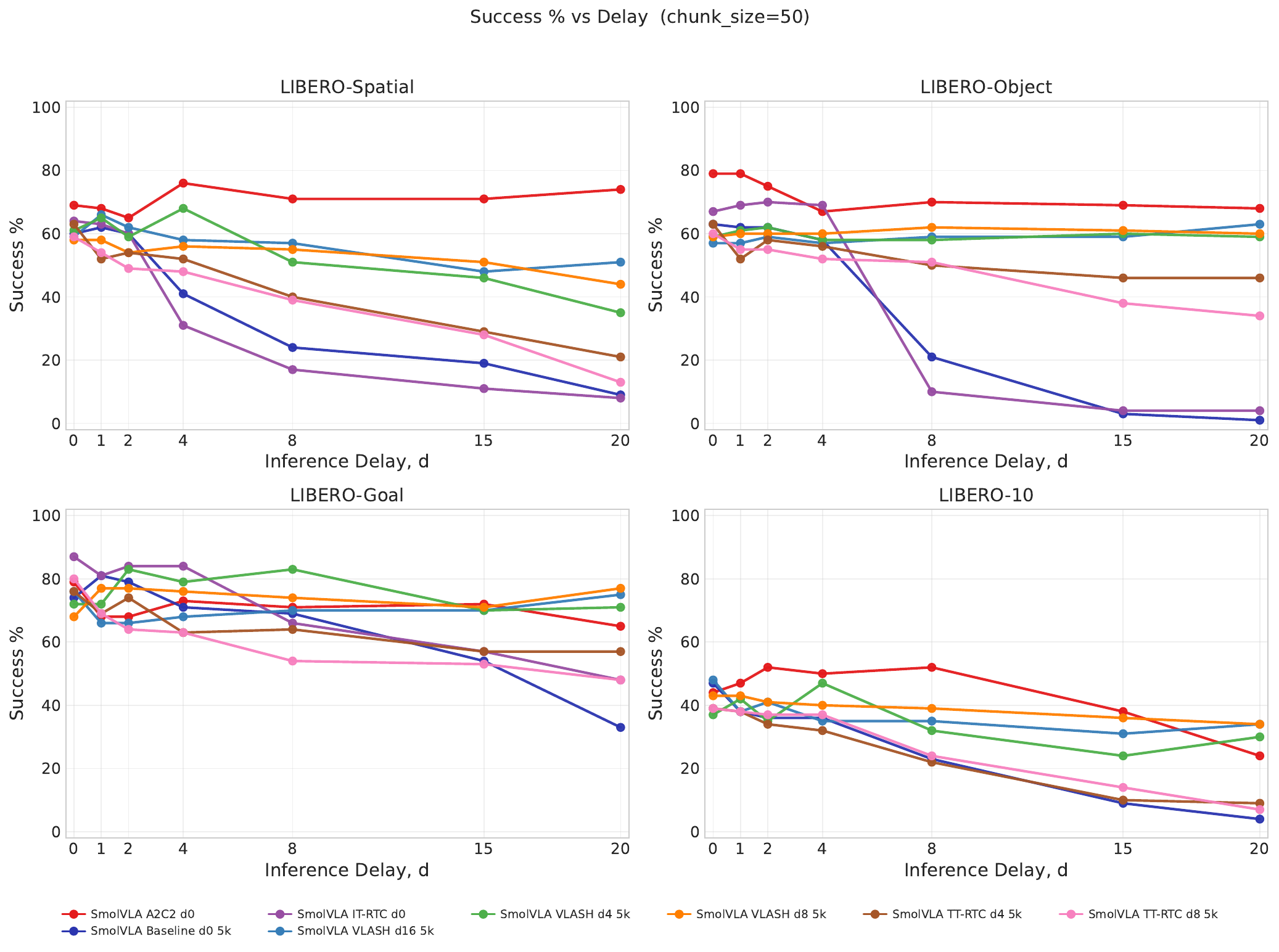}}
    \caption{Per-suite LIBERO success rate vs.\ delay ($s{=}50$).}
    \label{fig:libero_per_suite_success}
\end{figure}

\begin{figure}[H]
    \centering
    \includegraphics[width=0.95\textwidth]{\liberofig{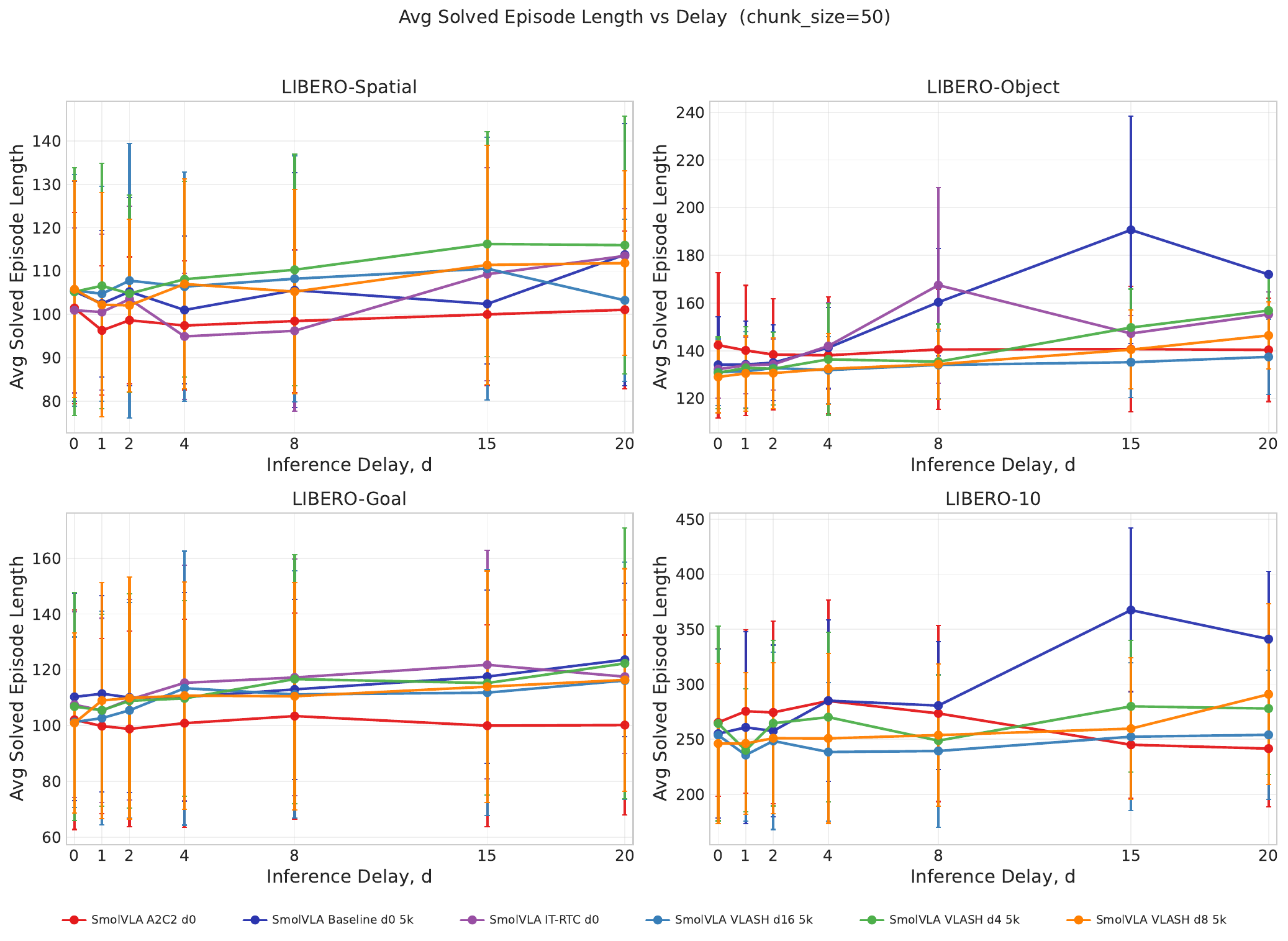}}
    \caption{Per-suite LIBERO solved-episode length vs.\ delay ($s{=}50$). Error bars: $\pm 1$ standard deviation.}
    \label{fig:libero_per_suite_ep_length}
\end{figure}

\section{VLASH Training-Configuration Ablation on LIBERO}\label{app:libero_vlash_ablation}

\begin{figure}[H]
    \centering
    \includegraphics[width=0.75\textwidth]{\liberoablationfig{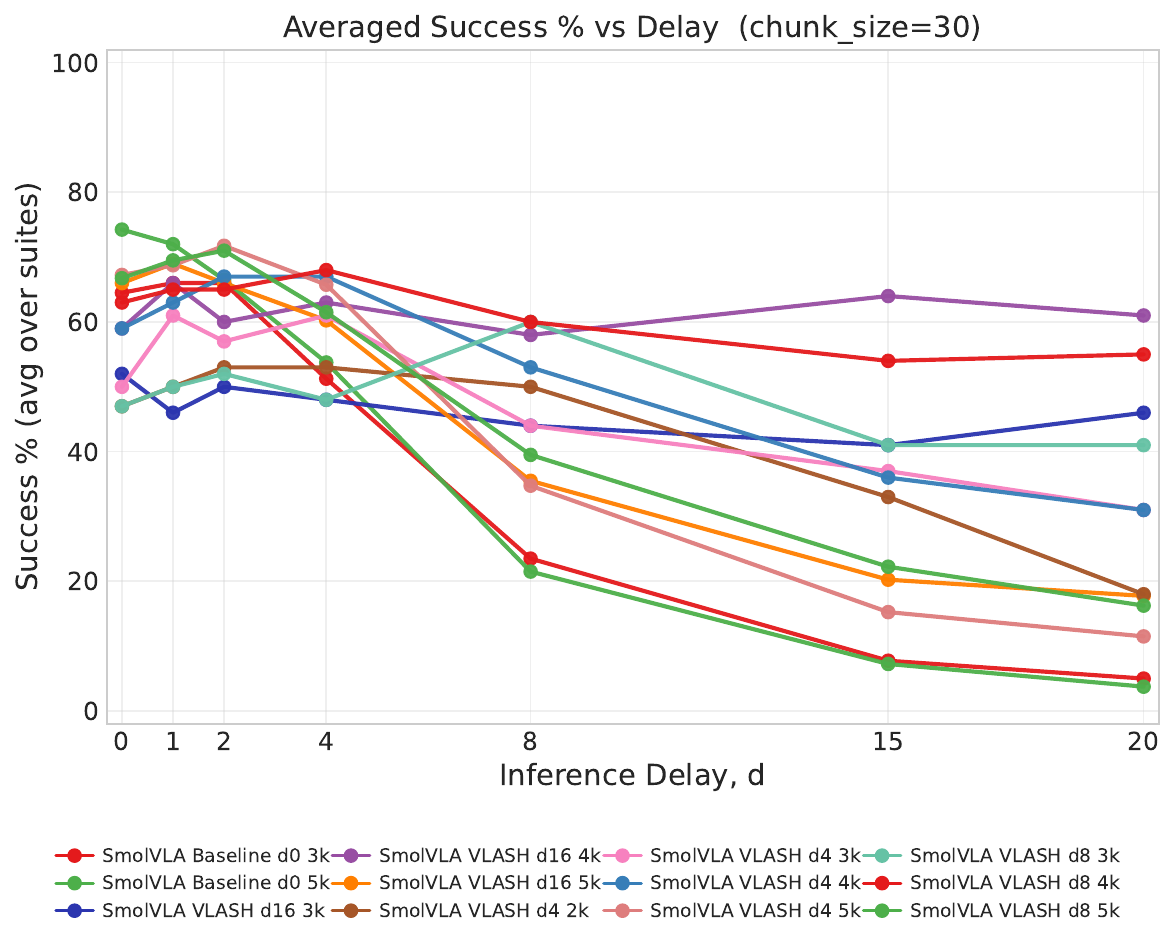}}
    \caption{VLASH training-configuration ablation on LIBERO: success rate vs.\ delay averaged across the four suites, for $d_{\max} \in \{0,4,8,16\}$ at multiple training checkpoints (2k--5k steps), evaluated with $s{=}30$.}
    \label{fig:libero_vlash_ablation}
\end{figure}

\end{document}